\pdfoutput=1
\documentclass[11pt,letterpaper]{article}
\usepackage[margin=1in]{geometry}
\usepackage{newtxtext}
\usepackage{authblk}
\usepackage{caption,tabularx,algorithm,algorithmic,placeins}
\usepackage{fancyhdr}
\date{}

\usepackage[T1]{fontenc}
\usepackage{cite}
\usepackage{amsmath,amssymb,amsthm,booktabs,array,url,graphicx}
\usepackage{newtxmath}
\usepackage[hidelinks]{hyperref}
\usepackage{microtype}
\newcommand{\FPR}{\operatorname{FPR}}
\newcommand{\BA}{\operatorname{BA}}

\newcommand{\cic}{CICIoT2023}
\newcommand{\nb}{N-BaIoT}
\newtheorem{theorem}{Theorem}

\newtheorem{corollary}{Corollary}


\newcommand{\TQSavingJ}{13.068}
\newcommand{\TQSavingJLow}{9.736}
\newcommand{\TQSavingJHigh}{16.400}

\newcommand{\TQFOneDropPP}{1.343}
\newcommand{\TQRecallDropPP}{2.954}

\newcommand{\DQSavingJ}{6.227}
\newcommand{\DQSavingJLow}{1.600}
\newcommand{\DQSavingJHigh}{10.853}

\begin{document}
\title{When Learned IDS Model Selection Collapses to a Static Choice: A Measurement Study at the Edge}
\author[1]{Nuonan Ouyang}
\author[2]{Bonan Zhang\thanks{Corresponding author: Bonan Zhang (\texttt{bnzhang@cityu.edu.mo}).}}
\author[1]{Adrian Shatte}
\author[3]{Zhigang Lu}
\author[4]{Chao Chen}
\author[5]{Wei Xiang}
\affil[1]{College of Science and Engineering, Division of Research, James Cook University, Bebegu Yumba Campus, Building 17, Douglas, Townsville, QLD, Australia}
\affil[2]{Faculty of Data Science, City University of Macau, Macau, China}
\affil[3]{School of Computer, Data and Mathematical Sciences, Western Sydney University, Penrith (Kingswood) Campus, Kingswood, NSW, Australia}
\affil[4]{College of Business and Law, RMIT University, Melbourne, VIC 3000, Australia}
\affil[5]{Australian Centre for AI in Medical Innovation and Cisco-La Trobe Centre for AI and Internet of Things, La Trobe University, Melbourne, VIC, Australia}
\maketitle
\begin{abstract}
Runtime model selection trades detection quality against energy and latency. However, comparing a selector only with a more costly static detector conflates detector choice with the benefit of adaptation. We present a framework for analysis and physical measurement that combines a pairwise reward criterion, matched static controls, paired blocks, a fixed load–prevalence factorial design, and reporting of actions outside training support. On one Raspberry Pi 4B with four frozen TON-IoT detectors, tabular Q-learning and a deep Q-network (DQN) both select LightLR throughout. A retrospective audit using validation data only rules out preference reversals for MedRF versus LightLR and HeavyMLP versus LightLR. However, TinyDT has a higher reward than LightLR in seven of the 2,500 windows, so the audit does not establish global LightLR optimality. We also evaluate a causal DeadlineGuard rule under a fixed factorial design with three arrival rates (50, 100, and 4,000 rows/s) and three attack prevalences (0.10, 0.50, and 0.90). Compared with Static-MedRF, DeadlineGuard saves 43.7 J per 500-s run (2.98\%; 95\% paired CI 41.4--46.0~J) and eliminates all 512 deadline misses among 5,000 windows, at a cost of 1.373 percentage points in F1. Outside training support, Tabular-Q's deterministic tie-breaking rule selects TinyDT, resulting in a false-positive rate of 67.76\%. These single-board results motivate joint audits of reward attainability, realized actions, attribution against matched static controls, service feasibility, and default actions outside training support.

\end{abstract}
\noindent\textbf{Keywords:}
Intrusion detection, IoT, edge resource management, model selection, reinforcement learning, service management, energy measurement, reproducibility.
\par\medskip
\section{Introduction}
Edge intrusion detection systems (IDSs) must balance security, latency and energy on tightly constrained hardware~\cite{Zhang2024VehicleEdge,Murti2024vRAN,Zhang2024DVFO,shu2026trustworthy}. Runtime model selection may help, but a learned controller may simply select the same low-cost detector throughout. Comparing it only with a more costly static detector then credits adaptation with what is really a change of operating point. Reward normalization can cause this before training starts, by making preference reversals unattainable. Sparse coverage of discrete states can also hide an unsafe default action until the workload moves outside the training support. 

We address these risks with an analysis and measurement framework. First, a pairwise non-degeneracy criterion tests whether two actions can exchange rank under the declared reward, cost profile, and validation support. Second, the measurement protocol includes five components: trace-matched physical blocks, a static control using the detector selected by the policy, a factorial design fixed before execution, joint reporting of raw and drift-adjusted endpoints, and action reporting both within and outside training support.  Third, we apply the framework on one Raspberry Pi 4B with four frozen TON-IoT detectors. We evaluate tabular Q-learning (Tabular-Q), a deep Q-network (DQN), a fixed confidence cascade, and a causal DeadlineGuard rule. Under the fixed factorial design with arrival rates of 50, 100, and 4,000 rows/s and attack prevalences of 0.10, 0.50, and 0.90, DeadlineGuard saves 43.7~J per 500-s run relative to Static-MedRF (2.98\%; 95\% paired CI 41.4--46.0~J). It also eliminates all 512 deadline misses among 5,000 windows, at a cost of 1.373 percentage points (pp) in F1. The learned selectors use LightLR throughout the observed support. Outside training support, Tabular-Q's deterministic tie-breaking selects TinyDT, giving a false-positive rate 67.76\%.

\begin{figure}[t]
\centering
\includegraphics[width=0.62\linewidth]{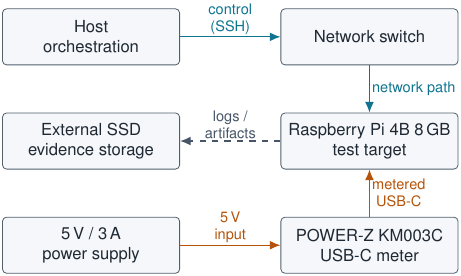}
\caption{Physical apparatus. A 5-V supply feeds the Raspberry Pi 4B through the POWER-Z KM003C USB-C meter; host orchestration reaches the Pi through the network switch; the external SSD stores evidence artifacts. Teal: network/control; amber: power measurement; dashed: evidence storage. The drawing shows roles and paths only.}
\label{fig:experimentdesign}
\end{figure}

The study addresses four research questions. \textbf{RQ1}: What security and resource operating points characterize the frozen detector pool? \textbf{RQ2}: Do learned selectors outperform the static schedule they realize, and does their reward allow preference reversals? \textbf{RQ3}: Do confidence- and deadline-aware controls provide useful switching under fixed and variable load? \textbf{RQ4}: How do state coverage, deadline violations, and threshold calibration using validation data only constrain deployment claims? We contribute a reproducible service-management method that distinguishes detector choice, conditional adaptation, objective design, and safety outside training support. We do not propose a new reinforcement learning (RL) algorithm. Fig.~\ref{fig:experimentdesign} shows the physical apparatus.

The remainder of the paper is organized as follows. Section~\ref{sec:relwork} reviews related work, and Section~\ref{sec:system} defines the system model and service objectives. Sections~\ref{sec:data}--\ref{sec:eval} describe the data, the detector pool, reward, and non-degeneracy criterion, and the evaluation protocol. Sections~\ref{sec:staticresults} and~\ref{sec:dynamicresults} report static/attribution and dynamic results, Section~\ref{sec:discuss} discusses implications, and Sections~\ref{sec:limits} and~\ref{sec:conclude} state limitations and conclude.

\section{Related Work}
\label{sec:relwork}
This work draws on edge IDS, conditional inference, model serving, and learned resource orchestration. Recent IDS research addresses continual detection, heterogeneous representations, and runtime model selection~\cite{Wang2024BFSNID,AOCIDS2024,StatAvg2025,DMIDS2025}. Auto-IDaS and FEDMS show that learned or confidence-driven detector selection is operationally plausible~\cite{Lin2024AutoIDaS,Hasan2026FEDMS}. However, neither controller complexity nor a lower-cost operating point alone establishes the value of adaptation. We therefore compare each selector with the static detector it actually uses and report the full state–action map alongside the occupancy of visited states.

Conditional-computation systems such as BranchyNet, adaptive network selection, MSDNet, and shallow–deep networks escalate uncertain inputs to costlier models~\cite{Teerapittayanon2016BranchyNet,Bolukbasi2017Adaptive,Huang2018MSDNet,Kaya2019ShallowDeep,wang2025medshield}. Our fixed confidence cascade applies this approach to whole-device IDS measurement. It also reveals a cost that row-level escalation rates can obscure. Only 1.34\% of rows are escalated, but these rows activate MedRF in 66.76\% of windows. Thus, call granularity, rather than the row fraction alone, governs the observed cost.

Related research also explores privacy-preserving neural architecture search and automated deep learning for secure inference~\cite{wang2022towards,wang2025privacy,shu2026trustworthy}. Serving systems such as Clipper, INFaaS, and Clockwork explicitly define service-level objectives (SLOs) for accuracy, latency, and cost, and route requests among model variants~\cite{Crankshaw2017Clipper,Romero2021INFaaS,Gujarati2020Clockwork}. We adopt this service-contract perspective and add two controls missing from these systems: a matched static action to separate the value of routing from model choice, and a pre-deployment audit to determine whether the declared reward allows preference reversals on validation support.

RL-based edge orchestration includes virtualized RAN (vRAN) reconfiguration, vehicle--edge scheduling, dynamic voltage and frequency scaling (DVFS), offloading, and QoS control~\cite{Murti2024vRAN,Zhang2024VehicleEdge,Zhang2024DVFO,Lou2024Startup,Kim2024DistributedOffload,Maleki2024QCS,Nan2026RobustDNN,Wang2025TrustOffload,Hu2025SITOff}. Tabular Q-learning and DQN are representative discrete and neural value-based controllers~\cite{Watkins1992,Mnih2015}. Our contribution to this line of work is a diagnostic methodology combining an analytic reward test, matched physical comparisons, a load–prevalence factorial design fixed before execution with an explicit deadline SLO, and an audit of actions outside training support. These components address known weaknesses in the evaluation of security ML and small-sample RL studies~\cite{Arp2022,Agarwal2021}.

\section{System Model and Service Objectives}
\label{sec:system}
An edge node processes batches in decision windows $t=1,2,\ldots$ using a resident detector pool $\mathcal M=\{m_1,\ldots,m_K\}$. Before inference, a frozen controller commits one action $a_t=\pi(s_t)\in\mathcal M$ from the causal state
\begin{equation}\label{eq:state}
s_t=(\widetilde T_t,h_{t-1},q_t,\lambda_{t-1},a_{t-1}),
\end{equation}
where $\widetilde T_t$ is smoothed temperature, $h_{t-1}$ the preceding selected detector's mean posterior, $q_t$ queue utilization, $\lambda_{t-1}$ preceding arrival count, and $a_{t-1}$ preceding action. Ground truth, current outputs of unselected detectors, and factorial-cell identity are unavailable at decision time. Because all models remain resident, selection excludes fitting and model-loading cost.

For each completed window we record attack-class F1, recall, and false-positive rate; completion latency is checked against deadline $\Delta_{\rm SLO}$; and resource cost is whole-device USB energy. In the physical study $\Delta_{\rm SLO}=1$~s and each run lasts 500~s. Training optimizes a scalar surrogate, but evaluation jointly reports security, timeliness, energy, action occupancy, and state support. Test policies are frozen and never updated online.

The network-layer adversary may vary traffic prevalence and arrival rate but cannot alter the meter, controller memory, detector artifacts, or training pipeline. Adversarial-ML attacks and hardware tampering are outside scope. The operational risk considered here is induced entry into unvisited discrete states: deterministic ties can then expose an unaudited default action.

We credit a selector with \emph{adaptive value} only when it realizes more than one action on deployment support, improves a service objective without a comparable regression, and survives a paired comparison with the static instance of its dominant realized action. Otherwise the measured difference is attributed to detector operating-point choice rather than conditional adaptation.

\section{Dataset-Native Data Preparation}
\label{sec:data}
\subsection{Choice of Primary Dataset}
The primary systems study uses TON-IoT~\cite{Moustafa2021TONIoT,TONSource}. Its records combine IoT service telemetry, operating-system events, and networked interactions labeled with attack categories, making the dataset representative of the mixed-signal traffic that reaches an edge IDS at the service boundary rather than at the wire only. We deliberately restrict the physical dynamic-power study to a single dataset because coupling multiple encoded pipelines to one hardware campaign would confound feature dimensionality, class balance, and physical measurement; instead \cic{}~\cite{Neto2023CICIoT2023,CICSource} and \nb{}~\cite{Meidan2018NBaIoT,NBaIoTSource} are retained as supplementary detector and coverage audits and are not used to imply cross-dataset dynamic-power generalization.

\subsection{Frozen Preprocessing and Splits}
The TON-IoT pipeline is frozen and retains 32 native features. Imputation, scaling, and categorical encoding are fitted on training data only and applied deterministically to validation and test. The split groups unordered IP pairs so that traffic patterns are not divided across partitions; the grouping rule hashes ``$11\vert$'' concatenated with the sorted IP pair and thresholds the first sixteen SHA-256 hexadecimal digits modulo 100 (training below 60, validation 60--79, test 80--99). Training seed 11 fixes all model fits, so any single-seed conclusion is explicit rather than implied. Source-row identifiers and raw-field-string fingerprints are cross-partition disjoint, and an independent encoded-float32 audit finds \emph{zero} TON-IoT train--test vector matches, which addresses a commonly cited leakage risk in IoT IDS benchmarks~\cite{Arp2022}.

\subsection{Auxiliary Dataset Audits}
The same ledger retains dataset-native \cic{} and \nb{} pipelines with 39 and 115 native features respectively. For \cic{}, seventeen attack labels remain training-only because their capture files are too few for a defensible test split; the actual test taxonomy is a binary attack-versus-normal partition over sixteen attack categories plus benign traffic. For \nb{}, physical device identifiers are the grouping unit, giving five training, two validation, and two test devices; a post-hoc byte-level audit records that 17.3\% of \nb{} test rows share exact float32 vectors with training rows (seven distinct vectors), reflecting the sensor-repetition structure of the dataset rather than a preprocessing fault. These coverage and encoded-overlap limitations are retained rather than used to imply cross-dataset dynamic-power generality; consequently the auxiliary datasets support only static classification and replay evidence.

\subsection{Class Balance and Replay Traces}
Class balance differs by dataset. Attacks are a minority of the TON-IoT test split (5{,}361 of 22{,}490 rows, 23.8\%) but a large majority of the \cic{} and \nb{} test splits (94.6\% and 89.8\%). We do not rebalance, because the physical study measures whole-device behavior on the natural distribution; prevalence dependence is instead made explicit through~\eqref{eq:prevalence} and joint reporting of F1, recall, and FPR. To probe operating points under controlled prevalence shifts, the supplement retains fifty static replay traces: ten stationary traces per dataset and ten prevalence-shift traces each for TON-IoT and \cic{}, whose successive phase attack fractions are $\{0.05, 0.35, 0.70, 0.40, 0.10\}$ over $\{100, 100, 120, 100, 80\}$ windows. Each trace is evaluated with all four detectors (200 model--trace evaluations). These traces do not constitute a new detector-training exercise; they resample fixed held-out pools to expose how a frozen detector's F1 and balanced accuracy behave when class prevalence changes while each class's feature distribution is held fixed.

\section{Detector Pool, Reward Specification, and Non-degeneracy Framework}
\label{sec:methods}
This section specifies the frozen detector artifacts, the reward specification used by both learned controllers, and the pre-deployment pairwise non-degeneracy screening test. The last of these is stated as a theorem so that it can be applied to future learned selectors without re-derivation.

\subsection{Detector Pool}
Each dataset has four separately fitted configurations: depth-6 TinyDT; L2-regularized LightLR; 25-tree, depth-10 MedRF; and a 64/32/16 HeavyMLP. All use threshold 0.5 and training seed 11. Detector names do not assert a resource ranking; the physical measurements determine the empirical order. During TON-IoT physical experiments all models remain resident, and an action selects one model for the next batch without loading, fitting, or updating it. This design deliberately isolates the effect of controller decisions from model instantiation cost so that any observed switching cost is attributable to inference itself, not to hot-swapping detectors.

\subsection{Causal State Encoding}
Before window $t$ the controller constructs $s_t$ per~\eqref{eq:state}. The kernel temperature is smoothed as $\widetilde T_t=0.9\widetilde T_{t-1}+0.1T_t$, initialized at $T_0$. The threat observation $h_{t-1}$ is the arithmetic mean attack probability from the single detector selected in the preceding window; it is \emph{not} the preceding ground-truth attack fraction. Queue utilization is bounded to $[0,1]$, and the preceding arrival count is divided by 100 and bounded to $[0,2]$. Initial values are $h_0=0.5$, $\lambda_0/100=1$, and $a_0=\text{TinyDT}$.

Each of the first four state variables has three bins. Their two boundaries are $(45,70)$~$^{\circ}$C, $(1/3,0.7)$ for threat, $(1/3,2/3)$ for queue utilization, and $(0.9,1.1)$ for arrival ratio. Including four previous actions yields $3^4\times 4=324$ encoded states. Boundary equality enters the higher bin. The flat index is row-major in the listed order, with previous action changing fastest.

The executed order within a window is: (i) state construction from causal telemetry and lagged context; (ii) action commitment and log; (iii) selected-detector inference; (iv) outcome log; and (v) lagged-state update for $t+1$. The information boundary of Section~\ref{sec:system} therefore holds at every decision; ground truth enters only offline training rewards, after the outcome is logged. This boundary is what allows preference-reversal analysis to be conducted purely on state features that the deployment can observe.

\subsection{Reward and Cost Provenance}
The unchanged per-window training reward is
\begin{equation}\label{eq:reward}
    r_t = 0.5\,B_t \;-\; 0.2\,\overline P_t \;-\; 0.2\,\overline L_t \;-\; 0.1\,\overline H_t,
\end{equation}
where $B_t$ is balanced accuracy over the classes present in the window. With $[x]_0^1=\min(1,\max(0,x))$, the normalization is
\begin{align}
    \overline P_t &= \biggl[\frac{P_t-2.347219336}{0.049806200}\biggr]_0^1,\\
    \overline L_t &= \biggl[\frac{L_t-1.2979595}{29.6547998}\biggr]_0^1,\\
    \overline H_t &= \biggl[\frac{\max(0,T_t^{\rm post}-\widetilde T_t)}{1\,^{\circ}\mathrm{C}}\biggr]_0^1.
\end{align}
$P_t$ and $L_t$ are frozen TON-IoT paced-profile lookups, not contemporaneous training measurements. Their means (power; batch latency) are 2.353132~W and 1.297960~ms for TinyDT, 2.347219~W and 1.544420~ms for LightLR, 2.397026~W and 30.952759~ms for MedRF, and 2.366688~W and 4.417836~ms for HeavyMLP. The compact software traces contain no post-action temperature response: their adapter supplies a fixed 35$^{\circ}$C predecision value and reuses that action-invariant value in the outcome field. Thus $\overline H_t=0$ throughout the executed training and validation traces; it is not an action-level thermal cost. No fitted thermal response or switching-energy penalty is used, and consequently no thermal control action is claimed.

\subsection{Pairwise Non-degeneracy Criterion}
Reward normalization is often reported as a hyperparameter without evidence that it permits any state to reverse the preferred action. We now show that this question can be answered before deployment with only the declared weights, cost profile, and a support region.

Consider two actions $a,b\in\mathcal M$. Decompose the scalar utility into a detection contribution and a normalized resource cost:
\begin{equation}
    U_a(s) = w_D\, D_a(s) - C_a,\qquad C_a=\sum_j w_j\,\overline C_{j,a},
\end{equation}
where $w_D=0.5$ in our specification and $\overline C_{j,a}$ collects the normalized resource penalties that are fixed on the declared support (as they are for the executed paced profiles at neutral thermal cost). Let $\Delta D_{ab}(s)=D_a(s)-D_b(s)$, $\Delta C_{ab}=C_a-C_b$, and $\tau_{ab}=\Delta C_{ab}/w_D$. Then
\begin{equation}\label{eq:diff}
    U_a(s)-U_b(s) = w_D\bigl(\Delta D_{ab}(s)-\tau_{ab}\bigr).
\end{equation}

\begin{theorem}[Pairwise non-degeneracy]\label{thm:nondeg}
Let $\mathcal S$ be a support region. Under the frozen scalar utility~\eqref{eq:diff}, a strict pairwise preference reversal between actions $a$ and $b$ exists on $\mathcal S$ if and only if
\begin{equation}\label{eq:nondeg}
    \min_{s\in\mathcal S}\Delta D_{ab}(s) < \tau_{ab} < \max_{s\in\mathcal S}\Delta D_{ab}(s).
\end{equation}
If \eqref{eq:nondeg} fails, $U_a(s)-U_b(s)$ cannot be strictly positive at one $s\in\mathcal S$ and strictly negative at another; consequently no policy that maximizes the frozen utility can strictly prefer $a$ on one state and $b$ on another within $\mathcal S$.
\end{theorem}

\begin{corollary}[Attainability under bounded detection]\label{cor:bound}
If $D_a(s),D_b(s)\in[0,1]$ for every $s\in\mathcal S$, then reversal is impossible whenever $|\tau_{ab}|\ge 1$; and reversal is guaranteed impossible whenever the attainable interval $[\min\Delta D_{ab},\max\Delta D_{ab}]$ does not straddle $\tau_{ab}$.
\end{corollary}

Because balanced accuracy lies in $[0,1]$, the corollary gives an immediate ceiling. Under the executed candidate-range scaling, the frozen nonthermal penalties imply $\tau_{\text{MedRF},\text{LightLR}}=0.7967$. A reproducible audit of all 2{,}500 windows in the five hash-bound validation traces finds $\Delta D_{\text{MedRF},\text{LightLR}}\in[-0.1053,0.1053]$, so this threshold lies outside the observed validation-window support. This retrospective audit excludes MedRF/LightLR and HeavyMLP/LightLR reversal on observed validation support. TinyDT/LightLR can reverse: TinyDT has higher immediate reward in 7 of 2{,}500 validation windows. These label-informed window comparisons establish neither optimality over all encoded states nor optimal long-horizon return, and do not fully explain the learned collapse. There is no evidence that this audit preceded test inspection. Supplementary Theorem~S1 gives the proof, Supplementary Table~S11A gives the full pairwise matrix, and Supplementary Algorithm~S1 gives the validation-only reward audit. If reversal is unattainable, the engineer should use an audited static schedule or renormalize against an external deployment contract---for example idle-relative power, the latency SLO, an energy envelope, or thermal headroom---rather than against detector-pool extrema.

\subsection{Policies}
Four policy families are evaluated in the primary physical campaign. \emph{Static-\{TinyDT, LightLR, MedRF, HeavyMLP\}} freeze a single detector. \emph{CFSM} is a fixed heuristic initialized to TinyDT; at a smoothed temperature of at least 70~$^{\circ}$C it steps down one model rank, bypassing cooldown; otherwise it holds for two windows after a switch and steps down/up one rank when threat crosses 0.7. Queue and load are logged but not used by CFSM. \emph{Tabular-Q} uses zero initialization, $\alpha=0.1$, $\gamma=0.9$, and the update
\begin{equation}
    \begin{aligned}
        Q(s_t,a_t) &\leftarrow Q(s_t,a_t) + 0.1\,\delta_t,\\
        \delta_t &= r_t + 0.9\max_a Q(s_{t+1},a) - Q(s_t,a_t),
    \end{aligned}
\end{equation}
with zero terminal bootstrap~\cite{Watkins1992}. \emph{DQN} uses one-hot state, two 64-unit ReLU layers, Adam $10^{-3}$, replay 10{,}000, batch 32, and a target copy every 250 updates~\cite{Mnih2015}. Both learners train for 1{,}000 episodes with $\varepsilon$ linearly annealed from 1 to 0.01; checkpoints are selected every 25 episodes on five validation traces. Test execution freezes the policy and performs no update.

Two prospectively frozen causal policies are added in the follow-up campaigns R1 and R2 (Section~\ref{sec:eval}). \emph{ConfidenceCascade} first executes LightLR on the full 100-row batch and, when its posterior $p_{\text{LR}}$ falls in $[0.3,0.7]$, invokes MedRF once on the deferred sub-batch. \emph{DeadlineGuard} uses only the current causal arrival count $N$: it selects MedRF when the predicted MedRF cost $\lceil N/100\rceil L_{95}^{\text{RF}}\le 1$~s, otherwise LightLR. Both thresholds were frozen from pre-experiment latency data.

Archived process-start and policy-creation timestamps indicate approximately 26.7~s for the Tabular-Q training run and 256.9~s for DQN; these are provenance-derived wall-clock intervals rather than calibrated compute benchmarks, and training energy was not instrumented. Deployment cost is measured separately in Section~\ref{sec:dynamicresults}.

\section{Evaluation Protocol and Measurement Instrumentation}
\label{sec:eval}
Our evaluation protocol is designed to separate detector choice from adaptive value along five instruments: (i) paired physical blocks across trace-matched detectors; (ii) matched-static-action attribution controls; (iii) prospectively frozen factorial workloads; (iv) separate raw and drift-adjusted power endpoints; and (v) full occupancy and off-support action reporting. This section describes each instrument together with the classification, timing, and power measurement pipelines that feed them.

\subsection{Classification and Static Evidence}
The held-out classification evaluation uses the frozen threshold 0.5 and reports attack-class F1, recall, balanced accuracy, false-positive rate, and the full confusion counts. In particular,
\begin{align}
    \FPR &= \frac{FP}{TN+FP},\qquad F1 = \frac{2TP}{2TP+FP+FN},\\
    \BA &= \tfrac12\!\left(\tfrac{TP}{TP+FN}+\tfrac{TN}{TN+FP}\right).
\end{align}
Counts are pooled over the declared evaluation unit before metrics are calculated. The supplement retains 50 static traces and 200 model--trace evaluations; these reuse fixed held-out pools and do not constitute fresh attacks, classifier retraining, or physical repeats.

\subsection{Pi Prediction Consistency and Inference Timing}
Pi parity checks cover 12 dataset--model combinations and 10{,}000 encoded rows each with \emph{zero} threshold-label disagreements against the training-host outputs. Tight-loop timing reports 100-row inference after warm-up and excludes encoding, transport, queueing, and controller cost; these boundaries are re-stated wherever a timing figure is used.

\subsection{Static Whole-Device Power}
The physical reference uses one Raspberry Pi 4B 8~GB and an inline POWER-Z KM003C USB-C meter. Fifteen valid 1{,}800-second segments comprise three repeats of instrumented idle and four TON-IoT detectors, all resident, at 100 pre-encoded rows/s. This is resource characterization, not another detection test.

The meter records nominally 1~Hz boundary samples, including raw ADC replies, signed current, voltage, and monotonic request/receive times. Under the recorded connector orientation, consumed power is $P_i=V_i(-I_i)$. Using the observed sampling times $\tau_i$, input energy is approximated by
\begin{equation}\label{eq:trap}
    E=\sum_{i=1}^{n-1}\frac{P_i+P_{i+1}}{2}(\tau_{i+1}-\tau_i),\qquad
    \bar P = \frac{E}{\tau_n-\tau_1}.
\end{equation}
No missing sample is filled. This is 1-Hz load-side USB energy, not calibrated wall-side measurement or a subsecond waveform.

\subsection{Statistical Units and Interval Reporting}
For an outcome $z$ observed over $n$ declared repetitions, the two-sided Student-$t$ interval is
\begin{equation}\label{eq:tci}
    \bar z\pm t_{0.975,\,n-1}\,\frac{s_z}{\sqrt n},
\end{equation}
using the sample SD~\cite{NISTMean}. Physical scheduler inference uses \emph{ten paired blocks}, never rows, windows, or meter frames. Raw energy is primary. Drift sensitivity subtracts each run's immediately preceding idle power; it is reported separately rather than substituted for raw measurement. Exact $2^{10}$ sign-flip tests supply a non-parametric robustness diagnostic under sign-symmetry and exchangeability, and are reported alongside the classical interval when both directional stability and magnitude precision matter.

\subsection{Physical Campaigns Overview}
Six campaigns follow the static characterization; four of them (primary, P1A, R1, and R2) are metered whole-device campaigns. The \emph{primary} campaign contains ten matched 500-second blocks of Static-MedRF, CFSM, Tabular-Q, and DQN. \emph{P1A} adds ten matched blocks with Static-LightLR to enable the attribution comparison against the static action the learned policies realize. \emph{P1B} isolates state-to-action decision timing on the Pi. \emph{P1C--P1E} are software retraining studies of seed, reward/state, and Tabular-Q $\alpha/\gamma$ variations. \emph{R1} adds ten matched blocks at 100 rows/s comparing Static-LightLR, the confidence cascade, frozen Tabular-Q, and Static-MedRF. \emph{R2} adds ten matched blocks and six methods over a prospectively frozen $50/100/4000\text{ rows/s}\times 0.10/0.50/0.90$ prevalence factorial with a fixed one-second service deadline. Two additional protocols are recorded: \emph{R3} was predeclared for controller-only power measurement but stopped before execution because its four-state design condition became ambiguous once the frozen artifact was inspected; \emph{R4} is a post-hoc TinyDT threshold diagnostic conducted on validation only. R3 and R4 are reported in full because omitting either would misrepresent the executed evaluation surface.

\section{Static and Attribution Results}
\label{sec:staticresults}
This section reports the classification operating points, the static physical resource characterization, and the primary/P1A attribution and robustness evidence. Together these anchor the pairwise non-degeneracy analysis of Section~\ref{sec:methods} in measured, rather than declared, cost profiles.

\subsection{TON-IoT Detection Operating Points}
Table~\ref{tab:classification} shows the four TON-IoT operating points used by the physical study. TinyDT has recall 0.9976 but F1 0.4792 and FPR 67.78\%; LightLR, MedRF, and HeavyMLP have F1 0.9620--0.9746 and FPR 1.35--1.52\%. MedRF's aggregate held-out balanced-accuracy advantage over LightLR is 0.0140; this post-freeze test statistic is descriptive and is not treated as a statewise attainability bound. The non-degeneracy conclusion instead uses the validation-only window audit in Section~\ref{sec:methods}, where the observed difference range $[-0.1053,0.1053]$ does not straddle $\tau_{\text{MedRF,LightLR}}=0.7967$. Full confusion counts, plus \cic{}/\nb{} detector and overlap audits, are retained in the supplement.

\begin{table}[t]
\centering\footnotesize
\caption{TON-IoT held-out detection at threshold 0.5. BA denotes balanced accuracy; FPR is a percentage.}
\label{tab:classification}
\setlength{\tabcolsep}{4.2pt}
\begin{tabular}{lrrrr}\toprule
Model & F1 & Recall & BA & FPR (\%)\\\midrule
TinyDT & 0.4792 & 0.9976 & 0.6599 & 67.78 \\
LightLR & 0.9620 & 0.9668 & 0.9766 & 1.35 \\
MedRF & 0.9746 & 0.9965 & 0.9906 & 1.52 \\
HeavyMLP & 0.9736 & 0.9903 & 0.9883 & 1.38 \\
 
\bottomrule\end{tabular}
\end{table}

For attack prevalence $p$, recall $r$, and FPR $f$,
\begin{equation}\label{eq:prevalence}
    F1 = \frac{2pr}{p(1+r) + (1-p)f},
\end{equation}
so high attack prevalence can coexist with high F1 and poor normal-class behavior. This identity motivates our decision to report attack-class F1, recall, and FPR jointly throughout, rather than summarizing with F1 alone.

\subsection{Static Physical Resource Measurements}
Across fifteen static segments, idle averages 2.335~W and active means span 2.347~W (LightLR) to 2.397~W (MedRF). MedRF's paired increment is 61.6~mW with 95\% CI $[32.1,\,91.0]$; TinyDT and LightLR intervals cross zero, indicating that the whole-device signal at this cadence is dominated by background platform power. No accepted workload window records undervoltage or throttling. Full segment statistics are supplementary; labels such as ``Tiny'' or ``Heavy'' are not used as empirical cost ranks.

\subsection{Primary Dynamic Comparison}
All forty primary runs contain 501 power samples and 500 decisions. Relative to Static-MedRF, Tabular-Q saves \TQSavingJ{} J (95\% paired CI \TQSavingJLow{}--\TQSavingJHigh{}) and DQN saves \DQSavingJ{} J (\DQSavingJLow{}--\DQSavingJHigh{}), but both lose \TQFOneDropPP{}~pp of F1 and \TQRecallDropPP{}~pp of recall. Both select LightLR in all 5{,}000 windows; consequently the primary contrast measures a detector operating-point change and not adaptive switching. This is the core observation that motivates the P1A attribution control below.

\subsection{P1A: Matched Attribution Against Static-LightLR}
P1A executes ten matched blocks comparing Tabular-Q, DQN, Static-LightLR, and Static-MedRF on the same traces. Because both learned policies realize LightLR throughout, Static-LightLR is the correct baseline for isolating adaptive value. Table~\ref{tab:p1a} reports three key paired contrasts.

\begin{table}[t]
\centering\footnotesize
\caption{P1A paired contrasts over ten trace-matched blocks. Each difference is first-minus-second method; intervals are Student-$t$ 95\% CIs, and $p$ is the exact two-sided sign-flip diagnostic over the ten block-level differences.}
\label{tab:p1a}
\setlength{\tabcolsep}{5pt}
\begin{tabular}{lrrr}\toprule
Outcome & Mean & 95\% CI & $p$\\\midrule
\multicolumn{4}{l}{\emph{Tabular-Q $-$ Static-LightLR}}\\
\quad Energy (J)  & $+0.456$  & $[-1.861,\,+2.772]$   & 0.664\\
\quad Power (mW)  & $+0.910$  & $[-3.722,\,+5.542]$   & 0.668\\
\multicolumn{4}{l}{\emph{DQN $-$ Static-LightLR}}\\
\quad Energy (J)  & $+4.525$  & $[+2.693,\,+6.357]$   & 0.00195\\
\quad Power (mW)  & $+9.045$  & $[+5.381,\,+12.709]$  & 0.00195\\
\multicolumn{4}{l}{\emph{Static-MedRF $-$ Static-LightLR}}\\
\quad Energy (J)  & $+12.471$ & $[+10.611,\,+14.331]$ & 0.00195\\
\quad F1 (pp)     & $+1.343$  & $[+1.292,\,+1.393]$   & 0.00195\\
\bottomrule\end{tabular}
\end{table}

Three findings emerge. First, Tabular-Q's energy contrast against the matched Static-LightLR control has an interval that includes zero, so the primary campaign's 13.1-J saving against Static-MedRF is fully explained by detector operating-point choice rather than by adaptation. Second, DQN carries a small but resolvable +4.5-J energy premium over Static-LightLR despite reproducing identical detection, indicating that its per-window controller cost, although a tiny fraction of each one-second window, is not free at 100 rows/s. Third, Static-MedRF's +12.5-J energy penalty against Static-LightLR is closely matched by the primary Tabular-Q saving against Static-MedRF, which confirms the compositional picture: (Static-MedRF $-$ Static-LightLR) $\approx$ (Static-MedRF $-$ Tabular-Q).

\subsection{P1B: Isolated Scheduler Overhead}
The isolated benchmark reports state-to-action decision time on the observed state support after correction of an earlier benchmarking substitution that was retained as an invalid attempt in the archive. On observed support, means are 0.739~$\mu$s for Static-LightLR, 6.582~$\mu$s for CFSM, 10.796~$\mu$s for Tabular-Q, and 1.567~ms for DQN. DQN is roughly $145\times$ Tabular-Q's controller cost but remains 0.157\% of the one-second decision interval. These timings are not converted into joules; the R2 physical evidence in Section~\ref{sec:dynamicresults} bounds what can be attributed to controller cost alone.

The full 240-unit benchmark, confidence intervals, and raw timing arrays are retained in Supplementary Table~S15.

\subsection{P1C--P1E: Multi-seed and Sensitivity Coverage}
P1C trains ten additional Tabular-Q and ten additional DQN policies under the frozen state, reward, split, and training-horizon protocol. All twenty policies select LightLR in 100\% of held-out windows (10/10 per algorithm; per-algorithm Wilson 95\% interval $[0.7225, 1.0]$). Greedy full-state maps vary across seeds---for DQN especially, from 0/324 to 311/324 TinyDT cells---but the observed-support behavior is unanimous, evidence for policy stability that is scoped to that support rather than to the whole encoded grid.

P1D varies the reward weights and separately ablates each state component; all 85 valid policies are single-action on held-out support. Only the five \emph{detection-only} Tabular-Q variants---those with all resource penalties removed---switch the collapsed action away from LightLR to MedRF, describing these frozen runs rather than proving that resource-aware policies must choose LightLR. All thirty state-ablation policies remain single-action LightLR, indicating that no single state component was the mechanism suppressing switching. P1E adds 35 Tabular-Q runs (five per setting) in one-at-a-time sweeps over $\gamma\in\{0,0.5,0.9,0.99\}$ and $\alpha\in\{0.05,0.10,0.20\}$; all thirty-five collapse to LightLR.

The full P1C--P1E matrices are retained in Supplementary Tables~S16--S18. Together with Table~\ref{tab:p1a}, they document the observed collapse through matched attribution, isolated timing, and sanctioned reward/state/hyperparameter perturbations. Only removing all resource penalties changes the collapsed detector in this matrix; this is not a proof of global optimality.

\section{Dynamic Scheduler Results Under Fixed and Variable Load}
\label{sec:dynamicresults}
\begin{table}[t]
\centering\footnotesize
\caption{R1 fixed-rate physical results; means over ten paired blocks. Adjusted power subtracts the immediately preceding idle. Escalation is the fraction sent from LightLR to MedRF.}
\label{tab:r1}
\setlength{\tabcolsep}{4.0pt}
\begin{tabular}{lrrrrrrr}\toprule
Method & Energy (J) & Power (W) & Adjusted (mW) & F1 & Recall (\%) & FPR (\%) & Escalation (\%)\\\midrule
Static-LightLR & 1360.4 & 2.721 & 37.0 & 0.9704 & 96.71 & 1.36 & -- \\
Cascade & 1368.3 & 2.737 & 50.6 & 0.9839 & 99.39 & 1.38 & 1.34 \\
Tabular-Q & 1362.0 & 2.724 & 39.7 & 0.9704 & 96.71 & 1.36 & -- \\
Static-MedRF & 1368.6 & 2.737 & 56.5 & 0.9839 & 99.66 & 1.54 & -- \\
 
\bottomrule\end{tabular}
\end{table}

\begin{table}[t]
\centering\footnotesize
\caption{R2 variable-load results over ten paired blocks. Energy is mean$\pm$SD across runs; deadline miss is the fraction of one-second windows missing the service deadline; completion p95 is the 95th-percentile window completion latency.}
\label{tab:r2}
\setlength{\tabcolsep}{3.1pt}
\begin{tabular}{lrrrrrrr}\toprule
Method & Energy (J) & Power (W) & F1 & Recall (\%) & FPR (\%) & Miss (\%) & Completion p95 (s)\\\midrule
Static-LightLR & 1416.8$\pm$3.7 & 2.834 & 0.9763 & 96.68 & 1.35 & 0.00 & 0.070 \\
Cascade & 1448.1$\pm$46.4 & 2.896 & 0.9902 & 99.41 & 1.36 & 0.42 & 0.665 \\
DeadlineGuard & 1424.4$\pm$5.2 & 2.849 & 0.9769 & 96.80 & 1.36 & 0.00 & 0.069 \\
Tabular-Q & 1400.9$\pm$48.0 & 2.802 & 0.7434 & 99.76 & 67.76 & 0.00 & 0.063 \\
DQN & 1461.1$\pm$123.9 & 2.922 & 0.9763 & 96.68 & 1.35 & 0.00 & 0.072 \\
Static-MedRF & 1468.1$\pm$4.8 & 2.936 & 0.9906 & 99.65 & 1.52 & 10.24 & 1.013 \\
 
\bottomrule\end{tabular}
\end{table}

\begin{figure}[!t]
\centering
\includegraphics[width=\textwidth]{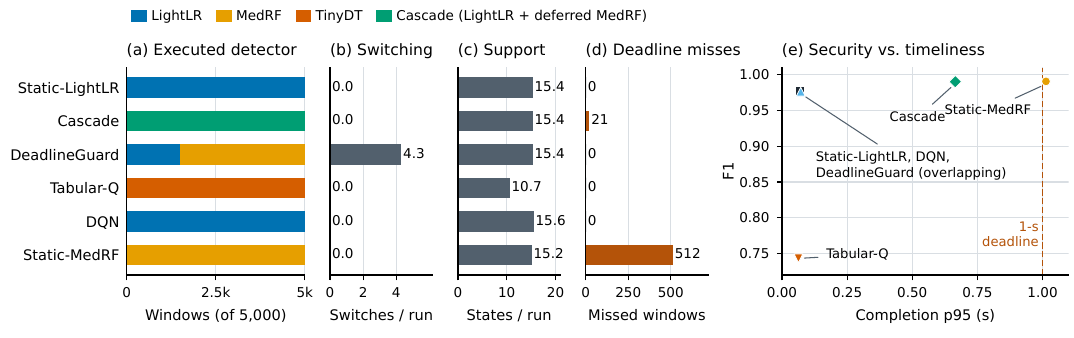}
\caption{R2 operational summary across the frozen 60-run factorial (six methods $\times$ ten blocks): (a) executed detector occupancy; (b) runtime switches per run, excluding the single initialization transition from the TinyDT placeholder $a_0$; (c) unique encoded states per run; (d) deadline misses; (e) F1 versus 95th-percentile window completion latency, with the 1-s deadline dashed. DeadlineGuard is the only method with load-conditioned switching, and it records zero deadline misses; Tabular-Q's TinyDT default under new load states appears in (a) and (e).}
\label{fig:r2dashboard}
\end{figure}

This section reports the two prospectively frozen follow-up campaigns and the two diagnostic protocols. R1 stresses a fixed-load confidence cascade; R2 stresses a load- and prevalence-varying factorial with an explicit one-second deadline SLO. R3 is a predeclared stop and R4 is a validation-only calibration diagnostic. Together they measure when conditional selection adds value and when it does not.

\subsection{R1: Fixed-Rate Confidence Cascade}
Table~\ref{tab:r1} reports ten new matched blocks at 100 rows/s. The fixed cascade escalates 1.34\% of rows and raises F1 by 1.347~pp over Static-LightLR while spending +7.952~J with 95\% CI $[5.897,\,10.007]$~J. Its raw mean-power difference is +15.917~mW $[11.798,\,20.035]$, whereas the preceding-idle adjustment gives +13.635~mW $[-1.515,\,28.784]$: environmental drift therefore weakens the power attribution but not the raw-energy result. Cascade F1 (0.98391) matches Static-MedRF (0.98386) within a few $10^{-5}$, at 0.279~pp lower recall and 0.156~pp lower FPR.

\subsubsection*{Cascade cost is a call-granularity phenomenon, not a bound}
The apparently disproportionate cascade latency at a 1.34\% escalation rate is explained by call granularity. The frozen source performs \emph{one} MedRF call per escalation-active window on the deferred sub-batch, not one MedRF call per deferred row. Although only 1.3388\% of the 500{,}000 rows are deferred, they land in 3{,}338 of 5{,}000 windows (66.76\%), with a mean of only 2.01 deferred rows per active window. No-escalation windows average 1.636~ms; any-escalation windows average 31.884~ms, close to the cost of a single MedRF batch call regardless of sub-batch size; the overall mean is 21.830~ms. This is an operating-point description, not a general bound: an optimized batched cascade with wider windows or amortized MedRF calls would show a different profile. What the measurement establishes is that reporting only the escalation \emph{fraction} would be misleading; escalation \emph{prevalence across windows} is the operational cost driver.

\subsection{R2: Variable-Load Factorial and DeadlineGuard}
R2 is the campaign in which conditional selection can, in principle, demonstrate adaptive value: every 500-second run visits each of the nine load-by-prevalence cells exactly once, followed by a 50-second recovery cell. All controllers see causal telemetry only; no method sees the factorial cell identity. Table~\ref{tab:r2} reports the six-method means; Table~\ref{tab:r2pairs} lists the two paired contrasts we treat as headline endpoints.

\begin{table}[t]
\centering\footnotesize
\caption{Headline R2 paired contrasts ($n=10$ blocks). Differences are first-minus-second method; F1 and deadline-miss changes are percentage points.}
\label{tab:r2pairs}
\setlength{\tabcolsep}{5pt}
\begin{tabular}{lrr}\toprule
 & DeadlineGuard $-$ & DQN $-$\\
Endpoint & Static-MedRF & Static-LightLR\\\midrule
$\Delta E$ (J)        & $-43.7$            & $+44.2$\\
\quad 95\% CI         & $[-46.0,\,-41.4]$  & $[-43.2,\,+131.7]$\\
$\Delta$F1 (pp)       & $-1.373$           & $+0.000$\\
$\Delta$Miss (pp)     & $-10.24$           & $+0.00$\\
Sign-flip $p$ (energy) & 0.00195           & 0.00195\\
 
\bottomrule\end{tabular}
\end{table}

\subsubsection*{DeadlineGuard: causal switching adds measurable value}
DeadlineGuard executes MedRF in 3{,}500 and LightLR in 1{,}500 of its 5{,}000 windows and makes 4.3 runtime switches per run on average (5.3 logged switches, one of which is the initialization transition from the TinyDT placeholder $a_0$ that the logs record for every policy whose first action is not TinyDT; Supplementary Table~S22). Against Static-MedRF, it uses 43.714~J less per run (95\% paired CI 41.418--46.010~J), a 2.98\% reduction relative to the 1{,}468.1-J Static-MedRF mean, while cutting the deadline-miss rate by 10.24~pp to zero. The trade-off is 1.373~pp lower F1 and 2.853~pp lower recall. All ten energy differences share sign (exact sign-flip $p=0.00195$). This is the campaign's positive result: a simple causal rule driven by a single observable, the current arrival count, achieves measurable energy savings and eliminates the SLO violations under variable load, an outcome that neither learned selector reproduces.

Against Static-LightLR, DeadlineGuard trades +7.550~J $[5.033,\,10.068]$ for a +0.06-pp F1 gain and preserves zero deadline misses. This second contrast is deliberately less dramatic: Static-LightLR is already feasible under the deadline SLO at 4{,}000 rows/s because its inference is fast; DeadlineGuard buys back a small detection-quality gap by escalating to MedRF only where that gap matters. The two contrasts together carry the design lesson: DeadlineGuard is not universally superior. Its benefit is large against the high-security static baseline that becomes infeasible at the deadline and marginal against the low-security static baseline that remains feasible.

\subsubsection*{The learned selectors fail to adapt under variable load}
Frozen DQN remains LightLR in all 5{,}000 windows and reproduces the LightLR detection endpoints. Frozen Tabular-Q, in contrast, reaches new load-related states never visited during training; the zero-initialized Q rows are tied, \texttt{np.argmax} returns the first maximum, and TinyDT is action index zero. The registry labels this ``fallback'' but the frozen source contains no separate exception branch: the R2 outcome is the hardware realization of the off-support tie behavior already visible in the 314/324-state TinyDT greedy map of the primary Q table. The resulting run-level F1 is 0.7434 and FPR is 67.76\%. This is a directly transferable safety risk of sparse discrete controllers whose default action ordering is not itself audited.

\subsubsection*{Heterogeneous variance and preceding-idle drift}
R2 raw energy is heterogeneous across methods. For DQN versus Static-LightLR, all ten raw block differences are positive (median +5.72~J; exact sign-flip $p=0.00195$), but a single Block-4 excursion is +392.0~J. That block's preceding idle is 3.919~W with a 49.2~$^{\circ}$C maximum temperature, unlike the other DQN blocks. The excursion inflates the SD of paired differences to 122.2~J and expands the Student-$t$ interval to $[-43.2,\,131.7]$~J. Preceding-idle adjustment does not resolve the DQN--LightLR contrast. We therefore retain both the consistent raw direction and the unstable magnitude and \emph{do not} claim a controller-power attribution, because R3 was not executed. This is exactly the sort of finding for which our protocol requires raw and drift-adjusted endpoints to be reported side by side. Figure~\ref{fig:r2dashboard} combines the corresponding occupancy, deadline, support, and security diagnostics.

\subsection{R3: Protocol-Preserving Stop}
R3 predeclared controller-only physical power over four ``original observed-support states'' for isolated attribution. The frozen artifact exposes eight such states and supplies no outcome-independent rule for selecting four. Choosing four \emph{after the fact} would allow controller-power reporting to be driven by known R2 results, so the campaign stopped before measurement per the predeclared stop rule. No controller-only joule value is claimed. Reporting the stop rather than filling it silently preserves the integrity of the pairwise DQN--LightLR contrast above, which is why the DQN R2 magnitude is deliberately left as unresolved.

\subsection{R4: Threshold Calibration Diagnostic}
R4 selects a TinyDT threshold on validation by maximum balanced accuracy (highest-threshold tie-break), hashes it, and applies it \emph{once} to test. Table~\ref{tab:r4} contrasts the frozen primary threshold with the validation-selected diagnostic.

\begin{table}[t]
\centering\footnotesize
\caption{TON-IoT TinyDT test results at the fixed primary threshold and the post-hoc validation-selected threshold. The diagnostic does not replace the primary threshold; it isolates calibration sensitivity as an explanation for the primary failure.}
\label{tab:r4}
\setlength{\tabcolsep}{5pt}
\begin{tabular}{lrr}\toprule
Quantity & Primary (fixed) & Validation-selected\\\midrule
Threshold $\tau$ & 0.500              & 0.646751\\
TN / FP          & 5{,}519 / 11{,}610 & 16{,}817 / 312\\
FN / TP          & 13 / 5{,}348       & 21 / 5{,}340\\
F1               & 0.4792             & 0.9698\\
Recall (\%)      & 99.76              & 99.61\\
FPR (\%)         & 67.78              & 1.82\\
BA               & 0.6599             & 0.9889\\
 
\bottomrule\end{tabular}
\end{table}

The diagnostic yields F1 0.9698, FPR 1.82\%, recall 99.61\%, BA 0.9889, and AUROC 0.9906. Recall changes by only 0.15~pp, so the primary failure at threshold 0.5 is substantially calibration-sensitive rather than a symptom of missing ranking information. Two consequences follow. First, TinyDT is not rehabilitated for the frozen physical campaigns: every classifier and scheduler artifact in those campaigns was fixed at threshold 0.5, and no post-hoc threshold can retroactively undo the unsafe TinyDT actions that Tabular-Q emitted under R2. Second, calibration governance---validation-only threshold selection with a hashed freeze before test---is a deployable practice that we recommend for any operational IDS pipeline that includes a TinyDT-class detector.

\section{Discussion}
\label{sec:discuss}
\subsection{Attribution and Conditional Value}
The primary 13.1-J Tabular-Q saving against Static-MedRF is only 26.1~mW over 500~s, while false negatives rise almost tenfold (57.8 to 565.9 per run). P1A resolves no Tabular-Q difference against the matched Static-LightLR action, so this is detector choice rather than adaptation. R2 supplies the positive contrast: DeadlineGuard uses the causal arrival count to switch, removes Static-MedRF's 512 deadline misses, and saves 43.7~J at a 1.373-pp F1 cost. The confidence cascade reaches MedRF-like F1, but its 1.34\% deferred rows activate MedRF in 66.76\% of windows; call granularity, not row fraction alone, controls serving cost.

\subsection{External-Budget Normalization Rule}
Theorem~\ref{thm:nondeg} makes reward design a deployment contract. Candidate-range scaling yields $\tau_{\text{MedRF,LightLR}}=0.7967$, outside the observed validation-window difference range $[-0.1053,0.1053]$, and therefore suppresses reversal on that support. A portable design replaces pool extrema with external denominators:
\begin{align}
\overline P_a&=\frac{P_a-P_{\rm idle}}{B_P}, &
\overline L_a&=\frac{L_a}{\Delta_{\rm SLO}},\\
\overline E_a&=\frac{E_a}{B_E}, &
\overline H_a&=\frac{[T_a-T_{\rm safe}]_+}{B_T},
\end{align}
where $B_P$, $\Delta_{\rm SLO}$, $B_E$, and $B_T$ are operator-set power, latency, energy, and thermal budgets. These denominators remain interpretable when the detector pool changes. The validation-only audit should be run before policy freeze; if no relevant pair can reverse, use an audited static schedule or redesign the pre-test contract.

\subsection{Support, Safety, and Measurement}
Frozen Tabular-Q maps 314/324 states to TinyDT because unseen zero-initialized rows tie and \texttt{np.argmax} selects action zero. R2 enters those new load states and raises FPR to 67.76\%; DQN remains LightLR, but seed-dependent global maps show that this is not validated off-support generalization. Deployments should therefore archive state coverage, full action maps, and the security quality of tied/default actions.

Small whole-device effects also require paired blocks and joint raw/drift-adjusted reporting. In R2, DQN-minus-LightLR is positive in all ten raw blocks, yet one thermal/power excursion widens the interval across zero; neither that direction nor the adjustment supports a controller-power claim because R3 stopped before measurement. The practical checklist is concise: audit reward attainability; include the realized static action and simple causal controls; freeze a load/threat stress factorial; report confusion counts, deadlines, occupancy, and off-support defaults; retain both raw and adjusted power endpoints; and select detector thresholds on validation only.

The portable contribution is this audit-and-attribution framework, not a universal numeric ranking. The 43.7-J DeadlineGuard saving, 66.76\% cascade activation, and controller timings are specific to the Pi 4B, resident pool, windowing regime, and meter chain used here.

\section{Limitations and Reproducibility Boundary}
\label{sec:limits}
Classifier results use one training seed, fixed dataset-native splits, and threshold 0.5; replay reuses held-out pools. Only TON-IoT receives dynamic physical testing. Ten paired blocks on one Raspberry Pi 4B 8~GB do not estimate board, meter, site, or fleet variability, and CICIoT2023/N-BaIoT are not dynamic-power evidence.

The ChargerLAB POWER-Z KM003C (HW~2.4, FW~2.0.6) was not independently laboratory calibrated. Its nominal 1-Hz load-side USB measurements exclude AC conversion, packet capture, and feature encoding. Models remain resident, temperatures stay low, and the factorial is scheduled feature replay rather than live traffic. The software-policy traces contain only a constant 35$^{\circ}$C predecision fallback and no action-attributable post-action temperature; the executed thermal reward is therefore zero and cannot establish thermal-aware adaptation. Pairing cannot remove ambient drift; R3 supplies no controller-only joule value; and R1 cascade cost is specific to 100-row windows.

P1 controls and R4 are post-hoc to the primary campaign; R1/R2 are prospectively frozen only for their own execution. The reward audit is retrospective, and cannot establish execution before test inspection. Its earlier implementation read aggregate test confusion counts for a descriptive diagnostic; the corrected entry point excludes those inputs. The corrected software adapter reports missing post-action temperature as null and flags the unavailable measurement; the thermal penalty is omitted rather than treated as a measured zero. Historical results are retained unchanged. A future temperature-increment experiment requires separate pre/post measurements on a common timing basis; a predecision EMA is a state feature, not a measured pre/post increment. No equivalence margin was declared. Invalid attempts and authorized replacements remain archived. Manifests, CRC, source hashes, and leakage validators establish integrity, not external validity. The theorem assumes a linear utility with $w_D>0$ and support-fixed normalized costs; non-linear objectives, switching penalties, and load-dependent costs require support-conditioned extensions.

\section{Conclusion}
\label{sec:conclude}
The pairwise non-degeneracy criterion tests action-pair reversal on a specified support region, while matched-static attribution distinguishes adaptation from detector choice. In our frozen study, Tabular-Q's apparent primary saving is explained by its LightLR operating point; off support its deterministic tie rule selects TinyDT and reaches 67.76\% FPR. By contrast, DeadlineGuard uses a direct causal load signal to save 43.7~J against Static-MedRF (2.98\%; 95\% CI 41.4--46.0~J) and remove all 512 deadline misses at a 1.373-pp F1 cost. Learned orchestration should be credited only after reward attainability, realized actions, matched controls, service feasibility, off-support defaults, and whole-device measurement are audited together.

\bibliographystyle{IEEEtran}
\bibliography{references}
\section*{Author information}
\begingroup\footnotesize
\noindent Nuonan Ouyang: \texttt{nuonan.ouyang@my.jcu.edu.au}; ORCID: 0009-0004-7124-8940.\\
Bonan Zhang (corresponding author): \texttt{bnzhang@cityu.edu.mo}; ORCID: 0000-0002-7394-7022.\\
Adrian Shatte: \texttt{adrian.shatte@jcu.edu.au}; ORCID: 0000-0002-6225-9697.\\
Zhigang Lu: \texttt{z.lu@westernsydney.edu.au}; ORCID: 0000-0001-5102-6217.\\
Chao Chen (Member, IEEE): \texttt{chao.chen@rmit.edu.au}; ORCID: 0000-0003-1355-3870.\\
Wei Xiang (Senior Member, IEEE): \texttt{w.xiang@latrobe.edu.au}; ORCID: 0000-0002-0608-065X.
\endgroup
\clearpage
\newgeometry{margin=22mm}
\begingroup
\fancyhead[L]{\small Supplementary Material}
\fancyhead[R]{\small IDS Model Selection at the Edge}
\setcounter{section}{0}
\renewcommand{\thesection}{S\arabic{section}}
\setcounter{theorem}{0}
\setcounter{corollary}{0}
\setcounter{figure}{0}
\setcounter{table}{0}
\renewcommand{\thetheorem}{S\arabic{theorem}}
\renewcommand{\thecorollary}{S\arabic{corollary}}
\renewcommand{\thealgorithm}{S\arabic{algorithm}}
\renewcommand{\thefigure}{S\arabic{figure}}
\renewcommand{\thetable}{S\arabic{table}}
\renewcommand{\theHsection}{supp.\arabic{section}}
\renewcommand{\theHtheorem}{supp.\arabic{theorem}}
\renewcommand{\theHcorollary}{supp.\arabic{corollary}}
\renewcommand{\theHfigure}{supp.\arabic{figure}}
\renewcommand{\theHtable}{supp.\arabic{table}}
\small
\setlength{\parindent}{0pt}
\setlength{\parskip}{5pt}
\renewcommand{\arraystretch}{1.12}

\begin{center}
{\LARGE Supplementary Material}\\[6pt]
{\large When Learned IDS Model Selection Collapses to a Static Choice:\\A Measurement Study at the Edge}\\[6pt]
Static and dynamic evidence, proofs, and audit procedure
\end{center}

\section{Scope and provenance}
Tables S1--S18 retain the accepted static, primary, and P0/P1 evidence. Tables S19 onward add the prospectively frozen follow-up controls R1/R2 and the post-hoc R4 diagnostic. R3 is retained as a predeclared protocol stop, not silently omitted. No classifier or frozen policy was retrained, and no primary result was overwritten.

The static-stage original large archive was not re-audited for this manuscript. Its retained independent-review artifacts remain the numerical source for Tables S1--S7; the current adaptive archive supplies the later sections. These evidence layers must not be conflated. The six-file executed-source reconciliation has since been closed by an archival-only V2 recovery package: all six historical source files match the runtime-bound SHA-256 values exactly, and the later source copies remain separately retained. The post-hoc P1 strengthening archive is now complete and is incorporated in the final sections below. The primary V1 campaign remains unchanged and retains its own evidence identity.

\section{Dataset grouping and evaluated coverage}
The retained split design differs by dataset. TON-IoT groups records by unordered IP pair; this is not a host-disjoint or chronological split. CICIoT2023 separates capture files within the grouping policy; categories with insufficient captures remain training-only. N-BaIoT separates physical-device identifiers into five training, two validation, and two test devices. Saved source-row identifiers and raw-field-string fingerprints are disjoint across partitions; this does not imply that all numeric encoded vectors are distinct.

CICIoT2023 has 34 original labels in the available classification taxonomy, comprising 33 attacks and normal traffic. The actual test coverage is \textbf{17 labels: 16 attack categories plus Benign\_Final}. Seventeen other attack categories are training-only. This is a binary attack-versus-normal evaluation, not a 34-class classification result.

\begin{table}[!htbp]
\centering\small
\caption*{\textbf{Table S1.} CICIoT2023 evaluated and training-only original labels. Columns are independent lists; rows do not denote matched categories.}
\begin{tabular}{ll}\toprule
Test labels & Training-only attack labels\\\midrule
Benign\_Final & Backdoor\_Malware \\
DDoS-ACK\_Fragmentation & BrowserHijacking \\
DDoS-ICMP\_Flood & CommandInjection \\
DDoS-ICMP\_Fragmentation & DDoS-HTTP\_Flood \\
DDoS-PSHACK\_FLOOD & DDoS-SlowLoris \\
DDoS-RSTFINFLOOD & DNS\_Spoofing \\
DDoS-SYN\_Flood & DictionaryBruteForce \\
DDoS-SynonymousIP\_Flood & DoS-HTTP\_Flood \\
DDoS-TCP\_Flood & MITM-ArpSpoofing \\
DDoS-UDP\_Flood & Recon-HostDiscovery \\
DDoS-UDP\_Fragmentation & Recon-OSScan \\
DoS-SYN\_Flood & Recon-PingSweep \\
DoS-TCP\_Flood & Recon-PortScan \\
DoS-UDP\_Flood & SqlInjection \\
Mirai-greeth\_flood & Uploading\_Attack \\
Mirai-greip\_flood & VulnerabilityScan \\
Mirai-udpplain & XSS \\
 
\bottomrule\end{tabular}
\end{table}

\FloatBarrier
\section{Device partition and encoded-vector diagnostic}
\begin{table}[!htbp]
\centering\small
\caption*{\textbf{Table S2.} N-BaIoT device allocation in the frozen partition. These are dataset device names, not additional physical test platforms in this study.}
\begin{tabular}{ll}\toprule
Partition & Dataset device identifier\\\midrule
Train & Ecobee\_Thermostat \\
Train & Ennio\_Doorbell \\
Train & Philips\_B120N10\_Baby\_Monitor \\
Train & Provision\_PT\_737E\_Security\_Camera \\
Train & SimpleHome\_XCS7\_1003\_WHT\_Security\_Camera \\
Validation & Danmini\_Doorbell \\
Validation & Samsung\_SNH\_1011\_N\_Webcam \\
Test & Provision\_PT\_838\_Security\_Camera \\
Test & SimpleHome\_XCS7\_1002\_WHT\_Security\_Camera \\
 
\bottomrule\end{tabular}
\end{table}

A post-hoc byte-level audit of saved float32 inputs identified 0 TON-IoT, 13 CICIoT2023, and 14,449 N-BaIoT test rows with an exact training-vector match. The N-BaIoT matches constitute 17.3343\% of its test rows and seven distinct encoded vectors. A further 21,832 N-BaIoT validation rows match training vectors. Distinct source-row identifiers and raw-field fingerprints can map to the same numeric input, including after preprocessing and float32 conversion. The matching vectors alone do not establish reuse of the same source row or the same physical device across partitions.

Table S3 excludes only test vectors matching \emph{training} inputs. Models, scores, and thresholds remain unchanged. The supplied CSV additionally retains the separate training-or-validation-match diagnostic, which is not substituted for Table S3 or the main result. All excluded N-BaIoT rows are attacks; its benign confusion counts and FPR are consequently unchanged. F1 changes also reflect the changed class mixture. Diagnostic stability does not eliminate all possible dataset bias.

\begin{table}[!htbp]
\centering\small
\caption*{\textbf{Table S3.} Post-hoc sensitivity excluding exact training-vector matches. Primary F1 uses the original test set; diagnostic F1, BA, and FPR use retained rows. BA denotes balanced accuracy; FPR is a percentage.}
\setlength{\tabcolsep}{3.4pt}
\begin{tabular}{llrrrrr}\toprule
Dataset & Model & Retained & Primary F1 & Diagnostic F1 & BA & FPR (\%)\\\midrule
TON-IoT & TinyDT & 22,490 & 0.479233 & 0.479233 & 0.659889 & 67.780 \\
 & LightLR & 22,490 & 0.961952 & 0.961952 & 0.976626 & 1.354 \\
 & MedRF & 22,490 & 0.974551 & 0.974551 & 0.990638 & 1.518 \\
 & HeavyMLP & 22,490 & 0.973593 & 0.973593 & 0.988261 & 1.378 \\
\midrule
CICIoT2023 & TinyDT & 71,132 & 0.990371 & 0.990369 & 0.829023 & 34.195 \\
 & LightLR & 71,132 & 0.987987 & 0.987985 & 0.786304 & 42.738 \\
 & MedRF & 71,132 & 0.991545 & 0.991544 & 0.850052 & 29.990 \\
 & HeavyMLP & 71,132 & 0.993015 & 0.993014 & 0.876306 & 24.739 \\
\midrule
N-BaIoT & TinyDT & 68,906 & 0.998373 & 0.997985 & 0.986099 & 2.764 \\
 & LightLR & 68,906 & 0.998393 & 0.998010 & 0.986632 & 2.646 \\
 & MedRF & 68,906 & 0.999733 & 0.999669 & 0.998095 & 0.366 \\
 & HeavyMLP & 68,906 & 0.997761 & 0.997228 & 0.980565 & 3.874 \\
 
\bottomrule\end{tabular}
\end{table}
The original held-out results remain the primary results. This analysis was conducted after inspecting the encoded inputs and must not be described as a preregistered alternative split, a model-selection procedure, or proof of complete input deduplication.

\FloatBarrier
\section{Static replay: corrected interval convention}
Each table entry summarizes ten replay seeds conditional on a fixed classifier, fixed preprocessing, and fixed held-out pool. The interval is
\[
\bar{x}\ \pm\ t_{0.975,9}\frac{s}{\sqrt{10}},\qquad t_{0.975,9}=2.2621571627.
\]
These are not ten independent classifier-training replications or ten physical experiments. The CSV retains unrounded values and all three corrected metric families: F1, balanced accuracy, and AUROC. The original mixture of normal-approximation output and Student-$t$ analysis is not carried into these tables.

\begin{table}[!htbp]
\centering\small
\caption*{\textbf{Table S4.} All twenty static replay F1 aggregates. The source scenario identifier \texttt{drift} is reported as \emph{prevalence-shift}, reflecting the actual manipulation rather than implying general concept-drift adaptation.}
\begin{tabular}{lllrrl}\toprule
Dataset & Replay & Model & Mean F1 & SD & Student-$t$ 95\% CI\\\midrule
CICIoT2023 & Prevalence-shift & HeavyMLP & 0.8090 & 0.0019 & [0.8077, 0.8104] \\
CICIoT2023 & Prevalence-shift & LightLR & 0.7104 & 0.0018 & [0.7091, 0.7117] \\
CICIoT2023 & Prevalence-shift & MedRF & 0.7778 & 0.0010 & [0.7770, 0.7785] \\
CICIoT2023 & Prevalence-shift & TinyDT & 0.7542 & 0.0010 & [0.7535, 0.7549] \\
CICIoT2023 & Stationary & HeavyMLP & 0.9931 & 0.0004 & [0.9928, 0.9934] \\
CICIoT2023 & Stationary & LightLR & 0.9881 & 0.0007 & [0.9876, 0.9887] \\
CICIoT2023 & Stationary & MedRF & 0.9917 & 0.0006 & [0.9913, 0.9921] \\
CICIoT2023 & Stationary & TinyDT & 0.9904 & 0.0007 & [0.9899, 0.9909] \\
N-BaIoT & Stationary & HeavyMLP & 0.9977 & 0.0001 & [0.9976, 0.9978] \\
N-BaIoT & Stationary & LightLR & 0.9984 & 0.0001 & [0.9983, 0.9984] \\
N-BaIoT & Stationary & MedRF & 0.9998 & 0.0001 & [0.9997, 0.9998] \\
N-BaIoT & Stationary & TinyDT & 0.9984 & 0.0002 & [0.9983, 0.9985] \\
TON-IoT & Prevalence-shift & HeavyMLP & 0.9822 & 0.0008 & [0.9816, 0.9828] \\
TON-IoT & Prevalence-shift & LightLR & 0.9704 & 0.0007 & [0.9699, 0.9710] \\
TON-IoT & Prevalence-shift & MedRF & 0.9839 & 0.0006 & [0.9834, 0.9843] \\
TON-IoT & Prevalence-shift & TinyDT & 0.6068 & 0.0010 & [0.6060, 0.6075] \\
TON-IoT & Stationary & HeavyMLP & 0.9733 & 0.0004 & [0.9730, 0.9737] \\
TON-IoT & Stationary & LightLR & 0.9616 & 0.0005 & [0.9612, 0.9619] \\
TON-IoT & Stationary & MedRF & 0.9744 & 0.0006 & [0.9740, 0.9748] \\
TON-IoT & Stationary & TinyDT & 0.4791 & 0.0010 & [0.4784, 0.4798] \\
 
\bottomrule\end{tabular}
\end{table}
Stationary traces contain 200 windows of 100 rows. Prevalence-shift traces contain 500 windows: successive phase lengths are 100, 100, 120, 100, and 80 windows, with attack proportions 0.05, 0.35, 0.70, 0.40, and 0.10. The replay seeds are 11, 23, 37, 53, 71, 89, 107, 131, 157, and 191. Prevalence-shift traces exist for TON-IoT and CICIoT2023, not for N-BaIoT in this evidence package. Each trace is evaluated with each fixed classifier, giving 200 completed static model--trace combinations.

The class-mixture schedule is not an online policy input. None of the entries in Table S4 establishes that a scheduler detected a phase change or adapted to an unknown attack. No adaptive policy was evaluated in this static stage.

\FloatBarrier
\section{Static replay: balanced accuracy}
\begin{table}[!htbp]
\centering\small
\caption*{\textbf{Table S5.} Balanced accuracy for the same twenty static replay aggregates. Intervals describe replay resampling under fixed trained artifacts.}
\begin{tabular}{lllrrl}\toprule
Dataset & Replay & Model & Mean BA & SD & Student-$t$ 95\% CI\\\midrule
CICIoT2023 & Prevalence-shift & HeavyMLP & 0.8762 & 0.0015 & [0.8752, 0.8773] \\
CICIoT2023 & Prevalence-shift & LightLR & 0.7862 & 0.0019 & [0.7849, 0.7875] \\
CICIoT2023 & Prevalence-shift & MedRF & 0.8502 & 0.0009 & [0.8495, 0.8508] \\
CICIoT2023 & Prevalence-shift & TinyDT & 0.8291 & 0.0009 & [0.8284, 0.8297] \\
CICIoT2023 & Stationary & HeavyMLP & 0.8773 & 0.0048 & [0.8739, 0.8807] \\
CICIoT2023 & Stationary & LightLR & 0.7870 & 0.0069 & [0.7820, 0.7920] \\
CICIoT2023 & Stationary & MedRF & 0.8518 & 0.0060 & [0.8474, 0.8561] \\
CICIoT2023 & Stationary & TinyDT & 0.8286 & 0.0058 & [0.8244, 0.8328] \\
N-BaIoT & Stationary & HeavyMLP & 0.9800 & 0.0011 & [0.9792, 0.9808] \\
N-BaIoT & Stationary & LightLR & 0.9862 & 0.0008 & [0.9856, 0.9868] \\
N-BaIoT & Stationary & MedRF & 0.9984 & 0.0005 & [0.9980, 0.9987] \\
N-BaIoT & Stationary & TinyDT & 0.9867 & 0.0012 & [0.9858, 0.9876] \\
TON-IoT & Prevalence-shift & HeavyMLP & 0.9883 & 0.0006 & [0.9879, 0.9888] \\
TON-IoT & Prevalence-shift & LightLR & 0.9767 & 0.0007 & [0.9762, 0.9772] \\
TON-IoT & Prevalence-shift & MedRF & 0.9906 & 0.0004 & [0.9903, 0.9909] \\
TON-IoT & Prevalence-shift & TinyDT & 0.6604 & 0.0015 & [0.6594, 0.6614] \\
TON-IoT & Stationary & HeavyMLP & 0.9882 & 0.0003 & [0.9880, 0.9884] \\
TON-IoT & Stationary & LightLR & 0.9765 & 0.0004 & [0.9762, 0.9768] \\
TON-IoT & Stationary & MedRF & 0.9906 & 0.0003 & [0.9904, 0.9908] \\
TON-IoT & Stationary & TinyDT & 0.6598 & 0.0005 & [0.6594, 0.6602] \\
 
\bottomrule\end{tabular}
\end{table}
The number of rows or windows is not an independent-replication count. These replay traces reuse a held-out pool and fixed predictions. Their confidence intervals do not incorporate training-seed variation, a different split, unseen devices, measurement calibration error, or a new environment. A bounded metric may yield a conventional unbounded Student-$t$ endpoint; such an interval is not silently clipped or replaced by a different convention. Full precision is retained in the data file.

\section{Uncertainty sources kept separate}
Classifier performance is conditional on training seed 11. The ten replay seeds describe resampling and phase composition conditional on the saved model and pool. Static physical power variability is based on three accepted repetitions per profile. The completed dynamic comparison has ten paired blocks, with four methods within each block; it does not provide forty independent paired differences for one method comparison. Training variability, replay variability, and physical repeatability must remain separately named.

An interval crossing zero does not prove equivalence. A statement that detection loss is ``small'' requires an explicitly justified margin and an honest account of when it was selected. A margin chosen after inspecting dynamic test outcomes must not be presented as prespecified.

\FloatBarrier
\section{Pi prediction agreement and tight-loop timing}
\begin{table}[!htbp]
\centering\small
\caption*{\textbf{Table S6.} Pi batch-inference timing. Each timing sample processes 100 already encoded rows; the tight loop follows twenty warm-up batches. The final column reports archived label disagreements in a separate 10,000-row parity check for each configuration, not a timing-sample count.}
\begin{tabular}{llrrrr}\toprule
Dataset & Model & Timing batches & Median (ms) & p95 (ms) & Label differences\\\midrule
TON-IoT & TinyDT & 1000 & 0.2448 & 0.3042 & 0 \\
 & LightLR & 1000 & 0.3306 & 0.3977 & 0 \\
 & MedRF & 1000 & 26.2880 & 27.4078 & 0 \\
 & HeavyMLP & 1000 & 1.5923 & 1.6725 & 0 \\
\midrule
CICIoT2023 & TinyDT & 1000 & 0.2323 & 0.2889 & 0 \\
 & LightLR & 1000 & 0.2999 & 0.3578 & 0 \\
 & MedRF & 1000 & 26.0360 & 27.8218 & 0 \\
 & HeavyMLP & 1000 & 1.5407 & 1.6281 & 0 \\
\midrule
N-BaIoT & TinyDT & 1000 & 0.2557 & 0.3142 & 0 \\
 & LightLR & 1000 & 0.3612 & 0.4275 & 0 \\
 & MedRF & 1000 & 26.2434 & 27.3492 & 0 \\
 & HeavyMLP & 1000 & 1.6570 & 1.7392 & 0 \\
 
\bottomrule\end{tabular}
\end{table}
The timing boundary excludes network capture, feature extraction, preprocessing, and the scheduling policy. These batch times must not be relabeled as single-record or end-to-end IDS latency. The power study used one batch per second; that pacing differs from the tight-loop timing study. The archived parity reports are not new hardware tests executed for this manuscript.

\section{Executed controller information boundary}
The frozen controller state consists of smoothed previous temperature, the previously selected detector's mean attack probability, lagged queue utilization, previous arrival count scaled by 100, and previous action. Initial temperature is read from the device; initial threat context is 0.5, initial load is 1, and the initialized action is TinyDT. The four numerical fields are discretized into three bins each and combined with four previous actions, giving 324 possible states. Main-text methods state the exact boundaries and reward.

Test labels and trace phase identities are not policy inputs. Current-window predictions are retrieved or computed only for the selected action. Independent reconstruction reproduces every formal action from the recorded decision-available state using the frozen Q table, DQN weights, or CFSM rule. A label-informed single-step reference is separately identified as non-deployable; it is not a global trajectory oracle.

\section{Recorded grouping function}
The recorded TON-IoT rule hashes \texttt{11|} followed by the sorted IP addresses joined with \texttt{|}; the integer value of the first sixteen SHA-256 hexadecimal digits modulo 100 assigns values below 60 to training, 60--79 to validation, and 80--99 to test. CICIoT2023 sorts capture paths by SHA-256 of \texttt{11|} plus relative path and allocates approximately 60/20/20, with at least one capture per partition when three or more exist. N-BaIoT similarly hashes \texttt{11|} plus device name and assigns the first five, next two, and final two devices. Exact retained memberships, rather than a reconstructed rounding rule, bind the evaluation.

\FloatBarrier
\section{Accepted static physical measurements}
\begin{table}[!htbp]
\centering\small
\caption*{\textbf{Table S7.} Fifteen accepted static segments. Duration and energy refer to the observed meter interval. Maximum temperature uses workload-window events only. Version identifies the retained execution/logging version, not a different classifier training.}
\begin{tabular}{rllrrrr}\toprule
Repeat & Profile & Version & Duration (s) & Energy (J) & Mean (W) & Max. ($^\circ$C)\\\midrule
1 & idle & v1 & 1800.001 & 4220.601 & 2.344777 & 33.102 \\
1 & TinyDT & v1 & 1799.997 & 4228.854 & 2.349367 & 33.102 \\
1 & HeavyMLP & v1 & 1800.004 & 4257.806 & 2.365442 & 34.076 \\
1 & LightLR & v1 & 1799.995 & 4224.380 & 2.346884 & 33.102 \\
1 & MedRF & v1 & 1800.002 & 4306.796 & 2.392662 & 33.589 \\
2 & idle & v1 & 1799.994 & 4194.960 & 2.330542 & 33.589 \\
2 & MedRF & v1 & 1799.996 & 4317.744 & 2.398752 & 33.102 \\
2 & LightLR & v1 & 1800.006 & 4226.906 & 2.348273 & 33.589 \\
2 & HeavyMLP & v1 & 1799.998 & 4248.284 & 2.360160 & 33.102 \\
2 & TinyDT & v2 & 1800.005 & 4244.018 & 2.357781 & 34.563 \\
3 & idle & v2 & 1799.995 & 4195.924 & 2.331076 & 33.102 \\
3 & LightLR & v3 & 1800.001 & 4223.704 & 2.346501 & 34.076 \\
3 & MedRF & v3 & 1799.998 & 4319.388 & 2.399662 & 35.050 \\
3 & HeavyMLP & v3 & 1800.000 & 4274.027 & 2.374460 & 35.537 \\
3 & TinyDT & v3 & 1800.000 & 4234.047 & 2.352248 & 36.024 \\
 
\bottomrule\end{tabular}
\end{table}
All segments use one Pi 4B 8 GB and TON-IoT classifiers, with four models resident. The idle profile retains the instrumented process and model residency. Each active profile replays 10,000 unique training rows eighteen times in 1,800 batches of 100 rows at one batch per second. Fifteen complete segments give 27,000 workload windows and 27,015 boundary-inclusive meter observations. Two interrupted attempts remain excluded and retained separately rather than spliced into complete runs.

Energy is obtained from observed power and actual adjacent timestamps using trapezoidal integration. Raw current signs are preserved in the source record; the recorded orientation is interpreted consistently when computing positive consumed power. No new static-stage raw-frame reintegration was performed: the saved independent-review segment calculations are the numerical source. Dynamic-stage raw frames, by contrast, were independently decoded and integrated for this manuscript. Repetition-level means, standard deviations, and matched-idle differences are recomputed from the retained segment records.

\section{Measurement and archive boundaries}
There is no independent laboratory calibration of the KM003C or independent ambient-temperature logger. Shared instrumentation and randomized paired ordering can improve comparability, but do not remove calibration uncertainty, environmental drift, carryover, or order effects. The low observed temperature range does not validate high-temperature operation or hard thermal guarantees. Whole-device USB-input power is not wall-plug AC energy, host energy, or a measured model-switching cost.

The historical root manifest has 371 valid listed entries. Two nested manifests were omitted:
\begin{quote}\small
\path{raw/materialization/MANIFEST_SHA256.txt}\\
\path{raw/materialization/materialization_run_v2/MANIFEST_SHA256.txt}
\end{quote}
A complete historical content list contains 373 files excluding the root manifest itself. The retained \path{CONTENT_SHA256_373.txt} refers to that original archive, not this smaller manuscript package. The manifest of this submission package is separate. The original archive is neither modified nor silently repackaged here.

\FloatBarrier
\section{Forty reconstructed physical runs}
Table S8 retains every formal attempt reported in the campaign. All forty contain 500 windows, 501 power observations, and 50,000 predicted rows. Actual integration durations are retained at full precision in \path{data/dynamic/run_registry_recomputed.csv}; they are approximately 500 seconds rather than replaced by that nominal constant. No failed formal run appears in the supplied execution log. The 120-second pilot is separate.

\begin{table}[!htbp]
\centering\small
\caption*{\textbf{Table S8.} Actual within-block order and reconstructed run-level outcomes. Mean batch latency is inference of 100 encoded rows in milliseconds; it excludes isolated controller decision overhead.}
\setlength{\tabcolsep}{3.2pt}\renewcommand{\arraystretch}{1.06}
\begin{tabular}{rrlrrrrrrr}\toprule
Block & Order & Method & Energy (J) & F1 & TN & FP & FN & TP & Batch (ms)\\\midrule
1 & 1 & Tabular-Q & 1192.491 & 0.969568 & 32342 & 458 & 585 & 16615 & 1.500 \\
1 & 2 & DQN & 1200.351 & 0.969568 & 32342 & 458 & 585 & 16615 & 1.450 \\
1 & 3 & Static-MedRF & 1210.268 & 0.983185 & 32282 & 518 & 68 & 17132 & 32.234 \\
1 & 4 & CFSM & 1193.732 & 0.636830 & 13435 & 19365 & 118 & 17082 & 2.345 \\
2 & 1 & DQN & 1197.694 & 0.971259 & 32336 & 464 & 523 & 16677 & 1.537 \\
2 & 2 & Static-MedRF & 1207.126 & 0.983275 & 32280 & 520 & 63 & 17137 & 32.637 \\
2 & 3 & Tabular-Q & 1191.666 & 0.971259 & 32336 & 464 & 523 & 16677 & 1.430 \\
2 & 4 & CFSM & 1196.817 & 0.634207 & 13239 & 19561 & 130 & 17070 & 2.398 \\
3 & 1 & Tabular-Q & 1190.721 & 0.969674 & 32350 & 450 & 589 & 16611 & 1.422 \\
3 & 2 & Static-MedRF & 1205.739 & 0.983306 & 32278 & 522 & 60 & 17140 & 32.618 \\
3 & 3 & CFSM & 1193.025 & 0.635446 & 13286 & 19514 & 103 & 17097 & 2.425 \\
3 & 4 & DQN & 1200.791 & 0.969674 & 32350 & 450 & 589 & 16611 & 1.464 \\
4 & 1 & Tabular-Q & 1190.126 & 0.969835 & 32361 & 439 & 594 & 16606 & 1.434 \\
4 & 2 & DQN & 1194.741 & 0.969835 & 32361 & 439 & 594 & 16606 & 1.451 \\
4 & 3 & Static-MedRF & 1209.233 & 0.984077 & 32295 & 505 & 50 & 17150 & 32.214 \\
4 & 4 & CFSM & 1195.999 & 0.634658 & 13245 & 19555 & 115 & 17085 & 2.361 \\
5 & 1 & Static-MedRF & 1206.460 & 0.983765 & 32285 & 515 & 51 & 17149 & 31.792 \\
5 & 2 & CFSM & 1197.197 & 0.637593 & 13465 & 19335 & 102 & 17098 & 2.284 \\
5 & 3 & DQN & 1201.542 & 0.969472 & 32345 & 455 & 591 & 16609 & 1.468 \\
5 & 4 & Tabular-Q & 1194.920 & 0.969472 & 32345 & 455 & 591 & 16609 & 1.494 \\
6 & 1 & Tabular-Q & 1191.543 & 0.971095 & 32398 & 402 & 587 & 16613 & 1.412 \\
6 & 2 & CFSM & 1199.311 & 0.639080 & 13626 & 19174 & 119 & 17081 & 2.316 \\
6 & 3 & DQN & 1207.523 & 0.971095 & 32398 & 402 & 587 & 16613 & 1.506 \\
6 & 4 & Static-MedRF & 1203.668 & 0.985005 & 32333 & 467 & 55 & 17145 & 32.358 \\
7 & 1 & CFSM & 1201.163 & 0.638346 & 13524 & 19276 & 100 & 17100 & 2.341 \\
7 & 2 & Tabular-Q & 1194.960 & 0.970849 & 32348 & 452 & 548 & 16652 & 1.427 \\
7 & 3 & DQN & 1195.812 & 0.970849 & 32348 & 452 & 548 & 16652 & 1.538 \\
7 & 4 & Static-MedRF & 1211.791 & 0.983611 & 32294 & 506 & 65 & 17135 & 33.472 \\
8 & 1 & CFSM & 1197.055 & 0.635367 & 13333 & 19467 & 128 & 17072 & 2.375 \\
8 & 2 & DQN & 1203.933 & 0.970564 & 32339 & 461 & 549 & 16651 & 1.441 \\
8 & 3 & Tabular-Q & 1197.662 & 0.970564 & 32339 & 461 & 549 & 16651 & 1.420 \\
8 & 4 & Static-MedRF & 1201.634 & 0.983505 & 32283 & 517 & 58 & 17142 & 33.466 \\
9 & 1 & Tabular-Q & 1195.749 & 0.971490 & 32359 & 441 & 537 & 16663 & 1.511 \\
9 & 2 & DQN & 1199.750 & 0.971490 & 32359 & 441 & 537 & 16663 & 1.518 \\
9 & 3 & Static-MedRF & 1205.626 & 0.984817 & 32315 & 485 & 44 & 17156 & 33.209 \\
9 & 4 & CFSM & 1199.494 & 0.638539 & 13553 & 19247 & 106 & 17094 & 2.278 \\
10 & 1 & CFSM & 1194.630 & 0.634338 & 13235 & 19565 & 123 & 17077 & 2.248 \\
10 & 2 & Tabular-Q & 1197.355 & 0.970581 & 32347 & 453 & 556 & 16644 & 1.420 \\
10 & 3 & Static-MedRF & 1206.329 & 0.984092 & 32310 & 490 & 64 & 17136 & 31.948 \\
10 & 4 & DQN & 1203.465 & 0.970581 & 32347 & 453 & 556 & 16644 & 1.444 \\
 
\bottomrule\end{tabular}
\end{table}

\FloatBarrier
\section{Full paired contrasts and diagnostic analysis}
\begin{table}[!htbp]
\centering\small
\caption*{\textbf{Table S9.} Method minus Static-MedRF, ten trace-matched differences per contrast. Detection differences are percentage points, not percentages relative to the baseline. The intervals are ordinary, unadjusted Student-$t$ 95\% intervals.}
\begin{tabular}{llrr}\toprule
Method & Outcome & Mean difference & 95\% CI\\\midrule
CFSM & Energy (J) & -9.9450 & [-12.7729, -7.1171] \\
CFSM & F1 (pp) & -34.7424 & [-34.8585, -34.6262] \\
CFSM & Recall (pp) & -0.3291 & [-0.3766, -0.2816] \\
CFSM & FPR (pp) & 57.6262 & [57.3297, 57.9227] \\
CFSM & BA (pp) & -28.9776 & [-29.1344, -28.8209] \\
CFSM & Mean batch (ms) & -30.2577 & [-30.6798, -29.8356] \\
CFSM & Max. temp. ($^\circ$C) & 0.1461 & [-0.1844, 0.4766] \\
Tabular-Q & Energy (J) & -13.0681 & [-16.4000, -9.7362] \\
Tabular-Q & F1 (pp) & -1.3425 & [-1.3926, -1.2924] \\
Tabular-Q & Recall (pp) & -2.9541 & [-3.0710, -2.8371] \\
Tabular-Q & FPR (pp) & -0.1738 & [-0.1964, -0.1512] \\
Tabular-Q & BA (pp) & -1.3901 & [-1.4425, -1.3378] \\
Tabular-Q & Mean batch (ms) & -31.1478 & [-31.5855, -30.7100] \\
Tabular-Q & Max. temp. ($^\circ$C) & -0.1948 & [-0.5693, 0.1797] \\
DQN & Energy (J) & -6.2270 & [-10.8535, -1.6005] \\
DQN & F1 (pp) & -1.3425 & [-1.3926, -1.2924] \\
DQN & Recall (pp) & -2.9541 & [-3.0710, -2.8371] \\
DQN & FPR (pp) & -0.1738 & [-0.1964, -0.1512] \\
DQN & BA (pp) & -1.3901 & [-1.4425, -1.3378] \\
DQN & Mean batch (ms) & -31.1131 & [-31.5350, -30.6911] \\
DQN & Max. temp. ($^\circ$C) & 0.1461 & [-0.2230, 0.5152] \\
 
\bottomrule\end{tabular}
\end{table}

The arithmetic mean of blockwise energy-saving percentages is 1.082170\% for Tabular-Q and 0.514887\% for DQN. Dividing the difference of method grand means by the Static-MedRF grand mean instead yields approximately 1.083\% and 0.516\%. Neither convention is substituted silently for the other. Absolute paired effects are the primary quantities.

\begin{table}[!htbp]
\centering\small
\caption*{\textbf{Table S10.} Exploratory order-adjusted energy contrasts (J), using categorical block, position, and method terms and block-clustered uncertainty. This post-hoc analysis does not replace the primary paired analysis or remove unmeasured environmental effects.}
\begin{tabular}{lrr}\toprule
Method & Adjusted difference & 95\% CI\\\midrule
CFSM & -9.542 & [-13.795, -5.289] \\
Tabular-Q & -12.523 & [-16.906, -8.139] \\
DQN & -6.243 & [-11.940, -0.547] \\
 
\bottomrule\end{tabular}
\end{table}
The supplied CSVs also preserve leave-one-block-out effects and exact two-sided sign-flip diagnostics over ten paired differences. The latter rely on the sign-symmetry/exchangeability assumptions of that diagnostic. A Holm correction across the three energy tests is explicitly post-hoc; it is not represented as a preregistered correction across all reported endpoints. None of these procedures incorporates instrument calibration uncertainty.

\FloatBarrier
\section{Action occupancy and detector selection}
\begin{table}[!htbp]
\centering\small
\caption*{\textbf{Table S11.} Detector occupancy (\% of 5,000 formal windows per method), mean true runtime switches, and mean initialization transitions per run. Initialization is not counted as runtime adaptation.}
\setlength{\tabcolsep}{4pt}
\begin{tabular}{lrrrrrr}\toprule
Method & TinyDT & LightLR & MedRF & HeavyMLP & Runtime & Initial\\\midrule
Static-MedRF & 0.00 & 0.00 & 100.00 & 0.00 & 0.0 & 1.0 \\
CFSM & 74.20 & 2.28 & 1.26 & 22.26 & 9.8 & 0.0 \\
Tabular-Q & 0.00 & 100.00 & 0.00 & 0.00 & 0.0 & 1.0 \\
DQN & 0.00 & 100.00 & 0.00 & 0.00 & 0.0 & 1.0 \\
 
\bottomrule\end{tabular}
\end{table}
\begin{figure}[!htbp]
\centering\includegraphics[width=.80\linewidth]{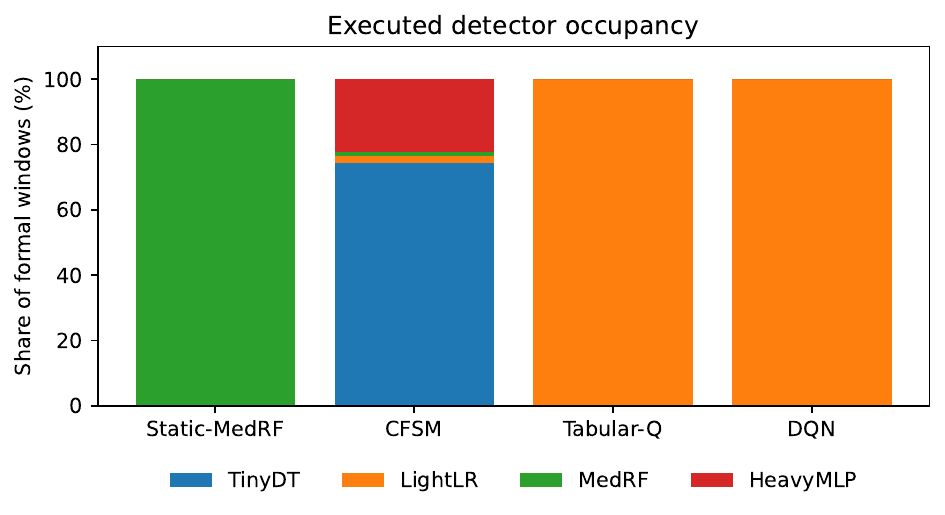}
\caption*{\textbf{Figure S1.} Both frozen learned policies execute a constant LightLR schedule. CFSM switches but spends most windows on the poorly performing TinyDT.}
\end{figure}
\begin{figure}[!htbp]
\centering\includegraphics[width=.80\linewidth]{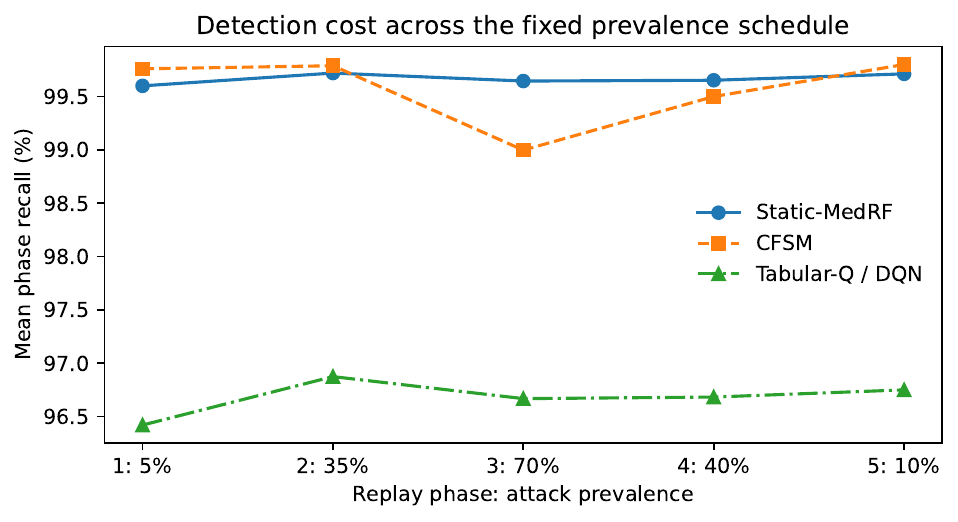}
\caption*{\textbf{Figure S2.} Mean recall by controlled prevalence phase. Tabular-Q and DQN produce identical selected predictions and are drawn as one series. These phases do not represent newly collected attacks or general concept drift.}
\end{figure}

\FloatBarrier
\section{Policy training and checkpoint selection}
Tabular-Q and DQN each use one policy-training seed, 20260907, with 1,000 software episodes of 500 windows. The ten training traces and five validation traces are source-ID disjoint from the ten test traces. Validation is recorded every 25 episodes. The frozen Tabular-Q checkpoint comes from episode 50 and the DQN checkpoint from episode 25 (one-based numbering), not automatically from the final episode. Full curves, checkpoint metadata, and all 324 reconstructed greedy state actions are retained. The original harness did not instrument training energy. Archived process-start timestamps (06:10:58 and 06:11:41 UTC) and policy-metadata creation timestamps (06:11:24.687905 and 06:15:57.922987 UTC) indicate approximately 26.7~s for the Tabular-Q run and 256.9~s (4.28~min) for DQN. These are provenance-derived wall-clock intervals rather than calibrated compute benchmarks.

\begin{figure}[!htbp]
\centering\includegraphics[width=.94\linewidth]{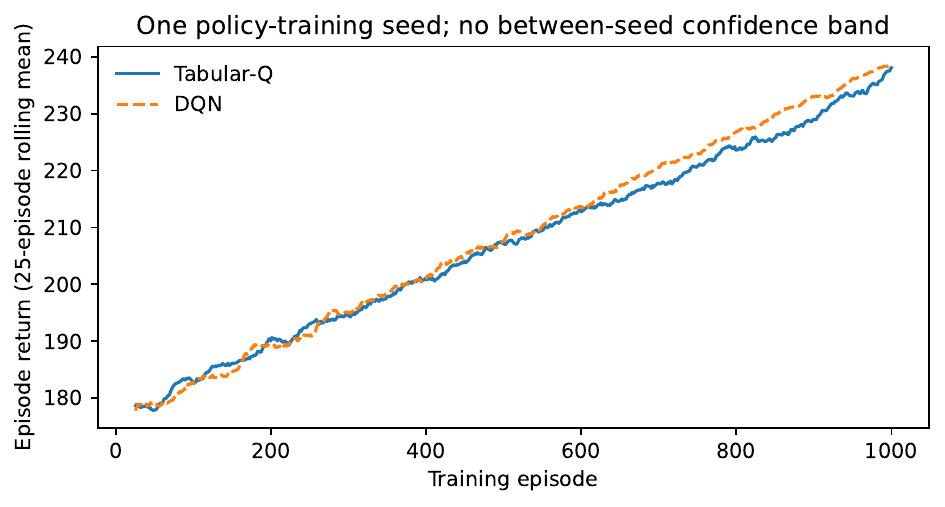}
\caption*{\textbf{Figure S3.} Training returns, displayed with a 25-episode trailing mean. Unsmoothed values are available in CSV. This is one realization per algorithm, not a multi-seed convergence guarantee.}
\end{figure}
\begin{figure}[!htbp]
\centering\includegraphics[width=.94\linewidth]{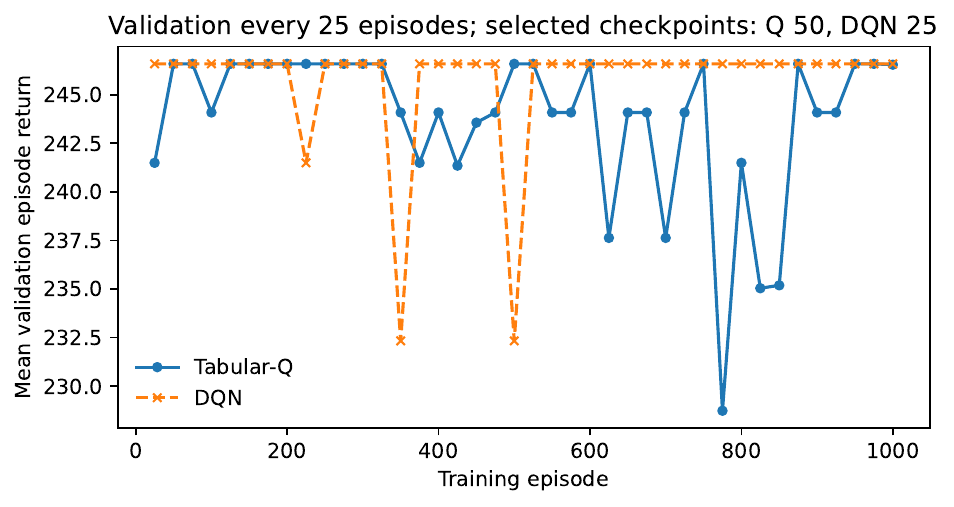}
\caption*{\textbf{Figure S4.} Recorded validation returns and checkpoint selection. The validation curve is not monotonic; the selected checkpoint is not evidence of convergence of every later iterate.}
\end{figure}

\FloatBarrier
\section{Post-hoc reward and state-coverage diagnostic}
\subsection{Pairwise non-degeneracy criterion}
For two actions $a,b$, write the frozen scalar utility difference as
\[
U_a(s)-U_b(s)=w_D\Delta D_{ab}(s)-\Delta C_{ab},\qquad w_D>0,
\]
where $\Delta C_{ab}$ is the normalized non-detection resource-penalty difference on the declared support, and define $\tau_{ab}=\Delta C_{ab}/w_D$. Then
\[
\operatorname{sign}(U_a(s)-U_b(s))=\operatorname{sign}(\Delta D_{ab}(s)-\tau_{ab}).
\]
\begin{theorem}[Pairwise non-degeneracy]\label{thm:supp-nondeg}
For $w_D>0$, a strict preference reversal between actions $a$ and $b$ exists on support $\mathcal S$ if and only if
\[
\min_{s\in\mathcal S}\Delta D_{ab}(s)<\tau_{ab}<\max_{s\in\mathcal S}\Delta D_{ab}(s).
\]
If this condition fails, $U_a(s)-U_b(s)$ cannot be strictly positive at one state of $\mathcal S$ and strictly negative at another.
\end{theorem}

\begin{proof}
Because $w_D>0$, the sign of $U_a(s)-U_b(s)$ equals the sign of $\Delta D_{ab}(s)-\tau_{ab}$. A strict reversal requires states $s,s'\in\mathcal S$ with opposite signs, equivalently one detection difference below $\tau_{ab}$ and another above it. This is exactly the strict interval condition in Theorem~\ref{thm:supp-nondeg}.
\end{proof}

\begin{corollary}[Bounded detection]
If $D_a(s),D_b(s)\in[0,1]$, reversal is impossible whenever $|\tau_{ab}|\ge 1$, or whenever the attainable interval $[\min\Delta D_{ab},\max\Delta D_{ab}]$ does not straddle $\tau_{ab}$.
\end{corollary}

With multiple normalized resource terms, $\Delta C_{ab}=\sum_j w_j(\bar C_{j,a}-\bar C_{j,b})$. Candidate-pool min--max scaling therefore makes the implied security--resource exchange rate depend on pool membership; an external power, latency-SLO, energy, or thermal budget provides a stable deployment reference.

\begin{algorithm}[!htbp]
\caption{Validation-only reward audit}\label{alg:supp-audit}
\footnotesize
\begin{algorithmic}[1]
\REQUIRE Actions $\mathcal M$; declared reward weights; cost profile $\{C_a\}$; validation support $\mathcal S$; detection scores $D_a$
\ENSURE Pairwise reversal-feasibility matrix $R$
\FOR{each unordered pair $(a,b)$}
  \STATE $\tau_{ab}\leftarrow(C_a-C_b)/w_D$
  \STATE $\Delta_{\min}\leftarrow\min_{s\in\mathcal S}\Delta D_{ab}(s)$; $\Delta_{\max}\leftarrow\max_{s\in\mathcal S}\Delta D_{ab}(s)$
  \STATE $R_{ab}\leftarrow\text{attainable}$ iff $\Delta_{\min}<\tau_{ab}<\Delta_{\max}$
\ENDFOR
\STATE Flag a reward as degenerate on $\mathcal S$ if no policy-relevant pair is attainable.
\end{algorithmic}
\end{algorithm}

Algorithm~\ref{alg:supp-audit} uses only quantities available at policy freeze and therefore does not inspect test outcomes. The corrected implementation reads only configuration, the trace manifest, and five validation traces, verifies their hashes, evaluates all four actions on each of 2{,}500 validation windows, and emits a pairwise CSV plus JSON audit. For MedRF minus LightLR it obtains $\tau=0.79667561$ and an observed validation-window balanced-accuracy difference range of $[-0.10526316,0.10526316]$, hence no strict reversal on that support. TinyDT exceeds LightLR in 7 of 2{,}500 validation windows; HeavyMLP never exceeds LightLR. The pairwise CSV reports reward-superiority counts and the JSON identifies all seven windows. This retrospective audit does not prove global LightLR optimality. Its previous entry point read aggregate test confusion counts for a descriptive diagnostic; those reads are removed. No evidence establishes execution before test inspection. The aggregate test difference 0.014011996 remains descriptive only. A failed audit calls for an audited static schedule or a pre-test redesign of the normalization, not post-hoc reward tuning.

\begin{table}[!htbp]
\centering\small
\caption*{\textbf{Table S11A.} Pairwise thresholds $\tau_{ab}$ under the frozen candidate-range normalization. Each value is the balanced-accuracy improvement action $a$ must offer over $b$ to overtake it.}
\setlength{\tabcolsep}{6pt}
\begin{tabular}{lrrrr}\toprule
$a\backslash b$ & TinyDT & LightLR & MedRF & HeavyMLP\\\midrule
TinyDT     & --       & $0.044$  & $-0.753$ & $-0.151$\\
LightLR    & $-0.044$ & --       & $-0.797$ & $-0.195$\\
MedRF      & $0.753$  & $0.797$  & --       & $0.602$\\
HeavyMLP   & $0.151$  & $0.195$  & $-0.602$ & --\\
\bottomrule\end{tabular}
\end{table}

\subsection{Executed reward decomposition}
The fixed reward is $0.5\,\mathrm{BA}-0.2\,\widetilde P-0.2\,\widetilde L-0.1\,\widetilde H$. The power min--max span is 0.049806200 W; the batch-latency span is 29.654800 ms. The compact software traces omit temperature telemetry, so the adapter supplies a constant 35 degrees Celsius predecision fallback and reuses that same, action-invariant value in the outcome field. The executed thermal penalty is consequently zero for every action; it is not a measured post-inference temperature rise and cannot provide action-level thermal attribution. The nonthermal resource penalty is 0.001662195 for LightLR and 0.4 for MedRF. Consequently, at equal thermal cost, MedRF requires an immediate balanced-accuracy improvement greater than 0.79667561 to exceed LightLR's single-window reward. This calculation explains the strength of the declared normalization; it is not a proof of a global optimal policy and did not lead to post-test reward changes.

Independent recomputation of the label-informed one-step reference selects LightLR in 4,991 of 5,000 held-out windows and HeavyMLP in nine, with neither MedRF nor TinyDT selected. This is a diagnostic reference using unavailable deployment labels, not a deployable comparison policy. Its CSV retains the objective and selected predictions.

All formal temperature and queue fields map to bin 0, and load maps to bin 1. Threat context varies, but there is no tested high-temperature or high-load region. The frozen DQN selects LightLR in every one of the 324 encoded states; the tabular policy selects LightLR in ten states and TinyDT in 314. Unvisited or zero-initialized tabular entries are not evidence of valid behavior outside the measured support. No thermal model was fitted or validated in this adaptive package, and no such performance is claimed.

\FloatBarrier
\section{Secondary pairwise and state-support diagnostics}
The primary analysis compares each adaptive method with Static-MedRF because that contrast was fixed in the original forty-run matrix. The following pairwise calculations are secondary diagnostics derived from the same ten blocks. They are useful for interpreting the realized policy but are not substituted for the primary endpoint.

\begin{table}[!htbp]
\centering\small
\caption*{\textbf{Table S12.} Selected secondary paired contrasts from the original campaign. Differences are first method minus second method; intervals use ten block-level Student-$t$ differences.}
\begin{tabular}{llrr}\toprule
Contrast & Metric & Mean difference & 95\% CI\\\midrule
Tabular-Q $-$ DQN & Energy (J) & -6.841 & [-9.717, -3.965] \\
Tabular-Q $-$ DQN & F1 & 0.000000 & [0.000000, 0.000000] \\
Tabular-Q $-$ DQN & Recall & 0.000000 & [0.000000, 0.000000] \\
Tabular-Q $-$ DQN & FPR & 0.000000 & [0.000000, 0.000000] \\
Tabular-Q $-$ CFSM & Energy (J) & -3.123 & [-5.461, -0.785] \\
Tabular-Q $-$ CFSM & F1 & 0.333998 & [0.332680, 0.335317] \\
Tabular-Q $-$ CFSM & Recall & -0.026250 & [-0.027565, -0.024935] \\
Tabular-Q $-$ CFSM & FPR & -0.578000 & [-0.580961, -0.575039] \\
 
\bottomrule\end{tabular}
\end{table}

Tabular-Q and DQN have identical selected predictions, so their detection differences are exactly zero rather than statistically estimated near-zero quantities. Tabular-Q nevertheless uses 6.841 J less energy on average than DQN (95\% CI 3.965--9.717 J in Tabular-Q's favor). Tabular-Q also uses 3.123 J less than CFSM (0.785--5.461 J) while producing much higher F1. These empirical mean relations do not identify isolated controller cost and do not generalize beyond the campaign.

\begin{table}[!htbp]
\centering\small
\caption*{\textbf{Table S13.} Greedy action map and observed state support for the two learned policies. ``All'' counts cover the 324 encoded states; ``Observed'' counts cover unique states reached during the formal learned-policy runs.}
\begin{tabular}{lrrrrrr}\toprule
Method & All states & Observed & All TinyDT & All LightLR & Obs. TinyDT & Obs. LightLR\\\midrule
Tabular-Q & 324 & 4 & 314 & 10 & 0 & 4 \\
DQN & 324 & 4 & 0 & 324 & 0 & 4 \\
 
\bottomrule\end{tabular}
\end{table}

Tabular-Q and DQN each reach only four unique states in formal execution. DQN maps every encoded state to LightLR. Tabular-Q maps ten encoded states to LightLR and 314 to TinyDT, but all four states actually reached by the learned-policy runs are LightLR states. Because most tabular states are never visited and many Q entries retain initialization structure, the all-state tabular map is not treated as validated deployment behavior outside observed support.

The phase-level diagnostic reaches the same conclusion. Neither learned policy changes action at a replay-phase boundary. Relative to Static-MedRF, mean learned-policy recall is lower by approximately 2.85--3.18 percentage points across each of the five phases while FPR is lower by approximately 0.16--0.20 percentage points. Phase-varying aggregate metrics therefore describe a constant LightLR operating point under changing class prevalence, not learned phase adaptation.

\section{Archive reconciliation and reproducibility limits}
All 558 top-level adaptive manifest entries verify. The runtime manifest separately references 57 artifacts: 22 match supplied bytes directly, 22 trace index/label arrays are independently reconstructed byte-exactly, seven depend on the frozen external baseline, and six source paths originally differed from the later source copies packaged with the adaptive ZIP. A self-consistent final ZIP and a byte-identical execution snapshot are therefore distinct verification questions.

The six runtime paths are \path{controller.py}, \path{worker.py}, \path{scheduler/artifacts.py}, \path{scheduler/cli.py}, \path{scheduler/evaluation.py}, and \path{scheduler/trace_io.py}. An archival-only V2 recovery package now contains all six historically executed byte sequences. Independent verification confirms 6/6 exact SHA-256 matches to the runtime-bound expected values, verifies all 17 content entries in its manifest, and retains the later source versions separately. No experiment was rerun and no V1 evidence was modified for this closure. The historical primary archive itself does not isolate scheduler decision overhead. That gap is addressed only by the separate P1B corrected microbenchmark described below; source recovery alone does not add measurement evidence.

\section{P1A matched Static-LightLR physical control}
P1A is a post-hoc attribution experiment designed after the primary learned policies were observed to select LightLR throughout. It contains ten matched blocks with Static-LightLR, Tabular-Q, DQN, and Static-MedRF. The archive retains 42 attempted runs: 40 valid formal runs, one DQN preparation failure before workload execution, and one Static-MedRF interruption after 403/500 power samples. The interrupted run is linked to exactly one authorized \texttt{retry1} replacement. No invalid attempt is silently deleted or combined with its replacement.

For every valid Tabular-Q, DQN, and Static-LightLR run, the selected-probability hash, confusion counts, and detection metrics are identical within the matched block. Each valid run contains 501 power observations and 500 decisions. Independent trapezoidal reintegration from the raw monotonic timestamps reproduces the registered run energy.

\begin{table}[!htbp]
\centering\footnotesize
\caption*{\textbf{Table S14.} P1A paired effects. Differences are first method minus second method. Exact sign-flip $p$ is a secondary robustness diagnostic; Student-$t$ intervals are the primary uncertainty summary.}
\setlength{\tabcolsep}{3.2pt}
\begin{tabular}{llrrr}\toprule
Contrast & Outcome & Mean & 95\% CI & Sign-flip $p$\\\midrule
Tabular--Q -- Static--LightLR & Energy (J) & 0.4555 & [-1.8609,2.7719] & 0.6641 \\
Tabular--Q -- Static--LightLR & Power (mW) & 0.9096 & [-3.7224,5.5416] & 0.6680 \\
Tabular--Q -- Static--LightLR & F1 (pp) & 0.0000 & [0.0000,0.0000] & 1.0000 \\
Tabular--Q -- Static--LightLR & Recall (pp) & 0.0000 & [0.0000,0.0000] & 1.0000 \\
Tabular--Q -- Static--LightLR & FPR (pp) & 0.0000 & [0.0000,0.0000] & 1.0000 \\
DQN -- Static--LightLR & Energy (J) & 4.5251 & [2.6933,6.3569] & 0.0020 \\
DQN -- Static--LightLR & Power (mW) & 9.0452 & [5.3812,12.7092] & 0.0020 \\
DQN -- Static--LightLR & F1 (pp) & 0.0000 & [0.0000,0.0000] & 1.0000 \\
DQN -- Static--LightLR & Recall (pp) & 0.0000 & [0.0000,0.0000] & 1.0000 \\
DQN -- Static--LightLR & FPR (pp) & 0.0000 & [0.0000,0.0000] & 1.0000 \\
Static--MedRF -- Static--LightLR & Energy (J) & 12.4706 & [10.6105,14.3306] & 0.0020 \\
Static--MedRF -- Static--LightLR & Power (mW) & 24.9412 & [21.2208,28.6615] & 0.0020 \\
Static--MedRF -- Static--LightLR & F1 (pp) & 1.3425 & [1.2924,1.3926] & 0.0020 \\
Static--MedRF -- Static--LightLR & Recall (pp) & 2.9541 & [2.8371,3.0710] & 0.0020 \\
Static--MedRF -- Static--LightLR & FPR (pp) & 0.1738 & [0.1512,0.1964] & 0.0020 \\
Tabular--Q -- DQN & Energy (J) & -4.0696 & [-6.4580,-1.6813] & 0.0098 \\
Tabular--Q -- DQN & Power (mW) & -8.1356 & [-12.9068,-3.3644] & 0.0098 \\
Tabular--Q -- DQN & F1 (pp) & 0.0000 & [0.0000,0.0000] & 1.0000 \\
Tabular--Q -- DQN & Recall (pp) & 0.0000 & [0.0000,0.0000] & 1.0000 \\
Tabular--Q -- DQN & FPR (pp) & 0.0000 & [0.0000,0.0000] & 1.0000 \\
 
\bottomrule\end{tabular}
\end{table}

The Tabular-Q minus Static-LightLR energy contrast is +0.4555 J with interval $[-1.8609,2.7719]$ J. Because no equivalence margin was predeclared, this is reported only as no resolved directional difference. DQN minus Static-LightLR is +4.5251 J with interval [2.6933, 6.3569] J. Static-MedRF minus Static-LightLR is +12.4706 J [10.6105, 14.3306] J and has higher detection quality. The supplementary campaign therefore shows that direct LightLR accounts for essentially the same detection operating point as both learned policies, while DQN carries an additional whole-device energy premium under the retained runtime.

\begin{figure}[!htbp]
\centering\includegraphics[width=.80\linewidth]{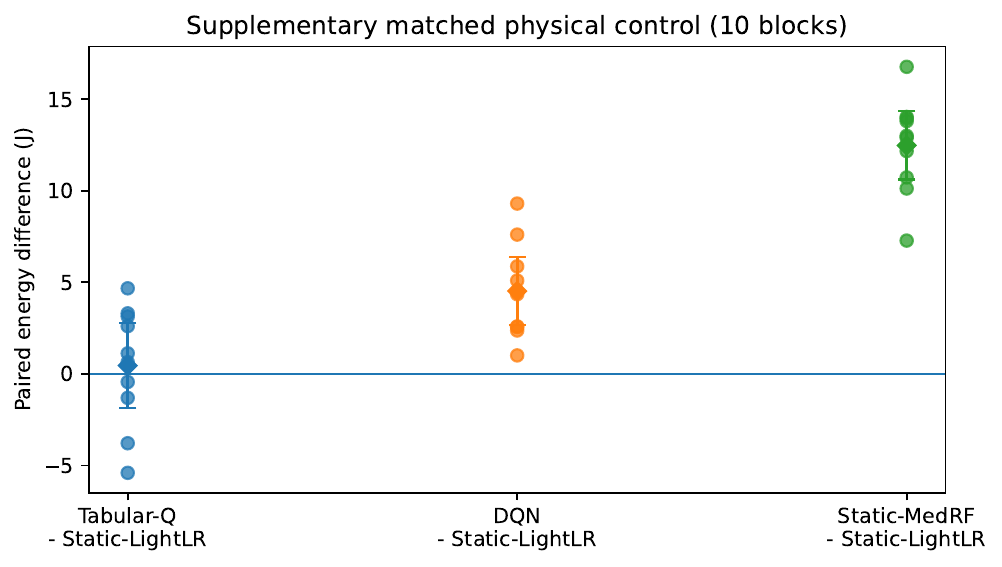}
\caption*{\textbf{Figure S5.} Ten P1A blockwise energy differences with Student-$t$ 95\% intervals.}
\end{figure}

\FloatBarrier
\section{P1B corrected isolated scheduler-overhead benchmark}
The first P1B attempt contained 240 timing units but failed the final validator because its CFSM path substituted fixed MedRF behavior and the required telemetry was absent. The complete attempt remains in the archive and is excluded from inference. A single authorized corrected replacement repeats the frozen benchmark design, adds telemetry, and passes 240/240 units. A deterministic 5,000-state comparison between the corrected CFSM and the frozen rule reports zero action mismatches.

Each method is measured in 30 paired benchmark blocks on each of two state streams: all 324 encoded states and the state support observed in the held-out traces. Each block executes 1,000 warm-up decisions and 10,000 timed state-to-action decisions. The raw nanosecond arrays reproduce the registered mean, median, p95, p99, and sample standard deviation exactly. The repeat unit is the benchmark block, not the 10,000 individual timings.

\begin{table}[!htbp]
\centering\footnotesize
\caption*{\textbf{Table S15.} P1B corrected decision overhead. The CI is across 30 block means; p95/p99 are means of block-level quantiles.}
\setlength{\tabcolsep}{3.3pt}
\begin{tabular}{llrrrr}\toprule
Stream & Method & Mean ($\mu$s) & 95\% CI & p95 ($\mu$s) & p99 ($\mu$s)\\\midrule
exhaustive\_324 & CFSM & 6.245 & 6.228--6.263 & 6.614 & 6.726 \\
exhaustive\_324 & DQN & 1559.645 & 1552.929--1566.362 & 1622.594 & 1651.166 \\
exhaustive\_324 & Static-LightLR & 0.634 & 0.626--0.641 & 0.689 & 0.842 \\
exhaustive\_324 & Tabular-Q & 10.443 & 10.420--10.466 & 10.643 & 10.947 \\
observed\_support & CFSM & 6.582 & 6.566--6.598 & 6.942 & 7.172 \\
observed\_support & DQN & 1566.897 & 1560.904--1572.890 & 1629.629 & 1668.873 \\
observed\_support & Static-LightLR & 0.739 & 0.716--0.761 & 0.952 & 1.127 \\
observed\_support & Tabular-Q & 10.796 & 10.764--10.828 & 11.047 & 11.286 \\
 
\bottomrule\end{tabular}
\end{table}

On observed support the means are 0.739~$\mu$s for Static-LightLR, 6.582~$\mu$s for CFSM, 10.796~$\mu$s for Tabular-Q, and 1.5669 ms for DQN. DQN is about 145 times the Tabular-Q state-to-action time but remains approximately 0.157\% of a one-second decision interval. These timings are not converted into energy.

\begin{figure}[!htbp]
\centering\includegraphics[width=.80\linewidth]{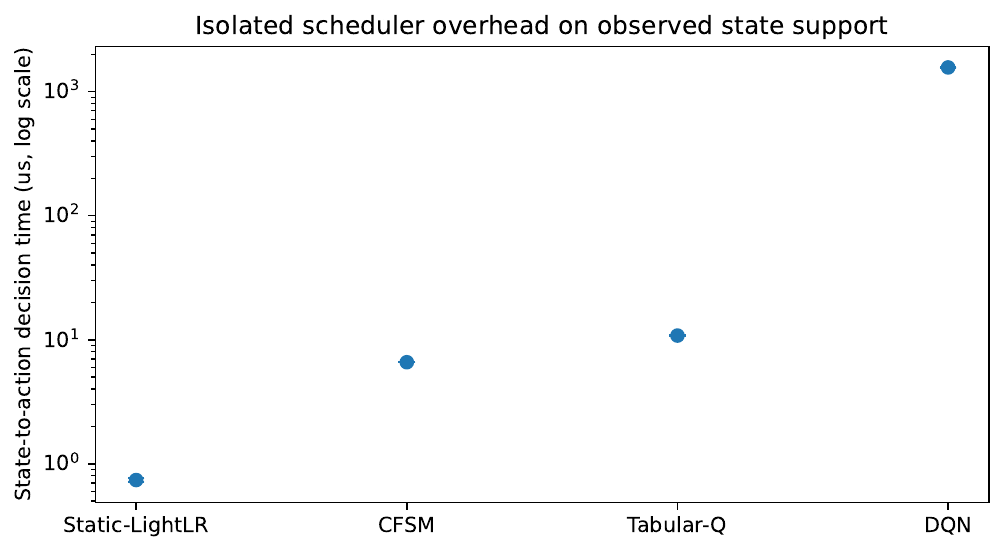}
\caption*{\textbf{Figure S6.} Corrected P1B state-to-action timing on observed state support. The vertical axis is logarithmic.}
\end{figure}

\FloatBarrier
\section{P1C multi-seed frozen-policy robustness}
P1C trains ten additional Tabular-Q and ten additional DQN policies under the frozen state, reward, data split, training horizon, and validation-only checkpoint rule. During final closure, the missing derived validation/test registry is produced from those already frozen checkpoints without retraining or checkpoint reselection. All twenty policies select LightLR in 100\% of held-out windows. Thus each algorithm has 10/10 held-out LightLR collapses, with Wilson 95\% interval [0.7225, 1.0000]. Raw switch counts include one initialization event per held-out trace; between-window switches are zero.

This robustness is support-limited rather than global. The action maps over all 324 encoded states vary across seeds, particularly for DQN. Table S16 reports the complete map counts while the final column reports held-out LightLR occupancy.

\begin{table}[!htbp]
\centering\footnotesize
\caption*{\textbf{Table S16.} P1C greedy global action maps and held-out occupancy. Counts cover all 324 encoded states.}
\setlength{\tabcolsep}{3.3pt}
\begin{tabular}{lrrrrrr}\toprule
Algorithm & Seed & TinyDT & LightLR & MedRF & HeavyMLP & Held-out LR (\%)\\\midrule
DQN & 1009 & 0 & 324 & 0 & 0 & 100.0 \\
DQN & 1013 & 205 & 119 & 0 & 0 & 100.0 \\
DQN & 1019 & 0 & 324 & 0 & 0 & 100.0 \\
DQN & 1021 & 0 & 324 & 0 & 0 & 100.0 \\
DQN & 1031 & 85 & 239 & 0 & 0 & 100.0 \\
DQN & 1033 & 192 & 10 & 0 & 122 & 100.0 \\
DQN & 1039 & 0 & 324 & 0 & 0 & 100.0 \\
DQN & 1049 & 311 & 13 & 0 & 0 & 100.0 \\
DQN & 1051 & 67 & 139 & 0 & 118 & 100.0 \\
DQN & 1061 & 0 & 324 & 0 & 0 & 100.0 \\
Tabular-Q & 1009 & 314 & 10 & 0 & 0 & 100.0 \\
Tabular-Q & 1013 & 313 & 11 & 0 & 0 & 100.0 \\
Tabular-Q & 1019 & 313 & 11 & 0 & 0 & 100.0 \\
Tabular-Q & 1021 & 312 & 12 & 0 & 0 & 100.0 \\
Tabular-Q & 1031 & 312 & 12 & 0 & 0 & 100.0 \\
Tabular-Q & 1033 & 314 & 10 & 0 & 0 & 100.0 \\
Tabular-Q & 1039 & 312 & 12 & 0 & 0 & 100.0 \\
Tabular-Q & 1049 & 312 & 12 & 0 & 0 & 100.0 \\
Tabular-Q & 1051 & 313 & 11 & 0 & 0 & 100.0 \\
Tabular-Q & 1061 & 313 & 11 & 0 & 0 & 100.0 \\
 
\bottomrule\end{tabular}
\end{table}

\FloatBarrier
\section{P1D reward/state sensitivity and P1E hyperparameter sensitivity}
P1D contains 85 valid policies plus one retained interrupted DQN attempt. Every valid policy is single-action on held-out support. Of the 85, 80 select LightLR. The only exceptions are the five Tabular-Q \texttt{detection\_only} reward policies, which select MedRF throughout and recover the MedRF-like F1. All thirty Tabular-Q state-ablation policies select LightLR. This supports specification sensitivity in a precise sense: the tested resource-aware objectives and state removals preserve the LightLR operating point, while deleting all resource penalties changes which static detector is preferred.

\begin{table}[!htbp]
\centering\footnotesize
\caption*{\textbf{Table S17.} P1D held-out policy results by family/variant. ``LR'' is the number of LightLR-collapsed policies; ``Single'' counts any single-action policy.}
\setlength{\tabcolsep}{3.2pt}
\begin{tabular}{lllrrrr}\toprule
Family & Algorithm & Variant & $n$ & LR & Single & Mean F1\\\midrule
Reward & DQN & default & 5 & 5/5 & 5/5 & 0.970439 \\
Reward & DQN & detection\_heavy & 5 & 5/5 & 5/5 & 0.970439 \\
Reward & DQN & energy\_heavy & 5 & 5/5 & 5/5 & 0.970439 \\
Reward & DQN & equal & 5 & 5/5 & 5/5 & 0.970439 \\
Reward & Tabular-Q & default & 5 & 5/5 & 5/5 & 0.970439 \\
Reward & Tabular-Q & detection\_heavy & 5 & 5/5 & 5/5 & 0.970439 \\
Reward & Tabular-Q & detection\_only & 5 & 0/5 & 5/5 & 0.983864 \\
Reward & Tabular-Q & energy\_heavy & 5 & 5/5 & 5/5 & 0.970439 \\
Reward & Tabular-Q & equal & 5 & 5/5 & 5/5 & 0.970439 \\
Reward & Tabular-Q & latency\_heavy & 5 & 5/5 & 5/5 & 0.970439 \\
Reward & Tabular-Q & thermal\_heavy & 5 & 5/5 & 5/5 & 0.970439 \\
State & Tabular-Q & full & 5 & 5/5 & 5/5 & 0.970439 \\
State & Tabular-Q & minus\_arrival & 5 & 5/5 & 5/5 & 0.970439 \\
State & Tabular-Q & minus\_previous\_action & 5 & 5/5 & 5/5 & 0.970439 \\
State & Tabular-Q & minus\_queue & 5 & 5/5 & 5/5 & 0.970439 \\
State & Tabular-Q & minus\_temperature & 5 & 5/5 & 5/5 & 0.970439 \\
State & Tabular-Q & minus\_threat & 5 & 5/5 & 5/5 & 0.970439 \\
 
\bottomrule\end{tabular}
\end{table}

P1E contains 35 valid Tabular-Q runs over the predeclared discount/learning-rate matrix. All 35 select LightLR throughout held-out support and are single-action. These results are sensitivity evidence only; none is used to choose a replacement primary hyperparameter after test inspection.

\begin{table}[!htbp]
\centering\footnotesize
\caption*{\textbf{Table S18.} P1E Tabular-Q hyperparameter sensitivity on held-out support.}
\begin{tabular}{lrrrr}\toprule
Variant & $n$ & LR & Single & Mean F1\\\midrule
alpha\_0.05 & 5 & 5/5 & 5/5 & 0.970439 \\
alpha\_0.10 & 5 & 5/5 & 5/5 & 0.970439 \\
alpha\_0.20 & 5 & 5/5 & 5/5 & 0.970439 \\
gamma\_0.00 & 5 & 5/5 & 5/5 & 0.970439 \\
gamma\_0.50 & 5 & 5/5 & 5/5 & 0.970439 \\
gamma\_0.90 & 5 & 5/5 & 5/5 & 0.970439 \\
gamma\_0.99 & 5 & 5/5 & 5/5 & 0.970439 \\
 
\bottomrule\end{tabular}
\end{table}

\FloatBarrier
\section{R1 fixed-rate confidence cascade}
R1 contains 40 valid 500-second runs in ten paired blocks. Static-LightLR, the frozen Tabular-Q policy, Static-MedRF, and the confidence cascade use the same trace within block; every method has an immediately preceding idle segment. The cascade band is fixed at $[0.3,0.7]$ and ground truth is never used for escalation. It sends 1.3388\% of rows to MedRF.

\begin{table}[!htbp]
\centering\footnotesize
\caption*{\textbf{Table S19.} R1 method means. Adjusted power subtracts the immediately preceding idle.}
\setlength{\tabcolsep}{3.0pt}
\begin{tabular}{lrrrrrrr}\toprule
Method & Energy & Power & Adjust. & F1 & Recall & FPR & Esc.\\
 & (J) & (W) & (mW) & & (\%) & (\%) & (\%)\\\midrule
Static-LightLR & 1360.4 & 2.721 & 37.0 & 0.9704 & 96.71 & 1.36 & -- \\
Cascade & 1368.3 & 2.737 & 50.6 & 0.9839 & 99.39 & 1.38 & 1.34 \\
Tabular-Q & 1362.0 & 2.724 & 39.7 & 0.9704 & 96.71 & 1.36 & -- \\
Static-MedRF & 1368.6 & 2.737 & 56.5 & 0.9839 & 99.66 & 1.54 & -- \\
 
\bottomrule\end{tabular}
\end{table}

Against Static-LightLR, paired cascade effects are +7.952 J [5.897,10.007], +15.917 raw mW [11.798,20.035], and +13.635 idle-adjusted mW $[-1.515,28.784]$. F1/recall/FPR differences are +1.347/+2.676/+0.018 percentage points. The raw and adjusted results are both retained because the adjusted interval indicates sensitivity to ambient/carryover drift.

The frozen worker implements sub-batch escalation, not one MedRF call per deferred row: it evaluates LightLR on the full 100-row batch and invokes MedRF once on the subset satisfying $0.3\le p_{LR}\le0.7$. The apparently disproportionate latency is explained by call granularity. Only 1.3388\% of rows are deferred, but they occur in 66.76\% of windows; among those windows the mean deferred subset has only 2.01 rows. A no-defer window averages 1.636 ms, whereas any-defer windows average 31.884 ms, nearly independent of whether one or several rows are deferred. Thus the 21.830-ms overall mean reflects frequent small MedRF calls at this 100-row cadence rather than row-by-row execution. It is an implementation/operating-point result, not a general bound for optimized cascades.

\begin{table}[!htbp]
\centering\footnotesize
\caption*{\textbf{Table S19a.} R1 confidence-cascade call-granularity diagnostic reconstructed from 5,000 frozen window logs.}
\begin{tabular}{lr}\toprule
Quantity & Value\\\midrule
Escalated rows / all rows & 6,694 / 500,000 (1.3388\%) \\
Windows with $\geq1$ escalated row & 3,338 / 5,000 (66.76\%) \\
Mean escalated rows per active window & 2.01 \\
No-escalation window inference & 1.636 ms \\
Escalation-active window inference & 31.884 ms \\
All cascade windows, mean inference & 21.830 ms \\
 
\bottomrule\end{tabular}
\end{table}

\begin{figure}[!htbp]
\centering\includegraphics[width=.95\textwidth]{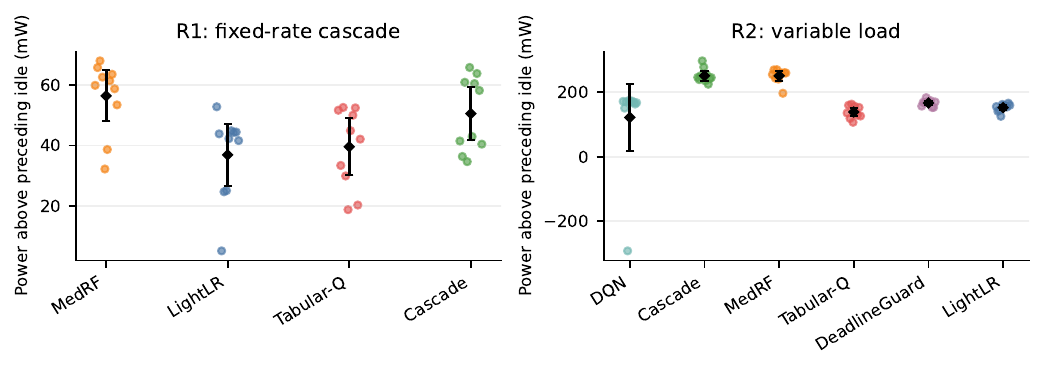}
\caption*{\textbf{Figure S7.} R1/R2 preceding-idle-adjusted power. Small points are runs; diamonds and bars are method means and Student-$t$ 95\% intervals over runs. Pairwise inference uses within-block differences in the CSV, not these marginal intervals.}
\end{figure}

\FloatBarrier
\section{R2 variable-load factorial and DeadlineGuard}
Each of the 60 valid runs visits every 50/100/4000 rows/s $\times$ 0.10/0.50/0.90 prevalence cell exactly once, then the fixed recovery cell. The controller sees causal measurements rather than cell identity. Static-MedRF misses 512 of 5,000 deadlines and Cascade misses 21; DeadlineGuard and the remaining methods miss none. All final backlogs are zero, so deadline misses and end-of-run backlog are distinct endpoints.

\begin{table}[!htbp]
\centering\footnotesize
\caption*{\textbf{Table S20.} R2 method results over ten blocks. Energy is mean$\pm$SD across physical runs; completion is seconds.}
\setlength{\tabcolsep}{2.6pt}
\begin{tabular}{lrrrrrrr}\toprule
Method & Energy & Power & F1 & Recall & FPR & Miss & p95\\
 & (J) & (W) & & (\%) & (\%) & (\%) & (s)\\\midrule
Static-LightLR & 1416.8$\pm$3.7 & 2.834 & 0.9763 & 96.68 & 1.35 & 0.00 & 0.070 \\
Cascade & 1448.1$\pm$46.4 & 2.896 & 0.9902 & 99.41 & 1.36 & 0.42 & 0.665 \\
DeadlineGuard & 1424.4$\pm$5.2 & 2.849 & 0.9769 & 96.80 & 1.36 & 0.00 & 0.069 \\
Tabular-Q & 1400.9$\pm$48.0 & 2.802 & 0.7434 & 99.76 & 67.76 & 0.00 & 0.063 \\
DQN & 1461.1$\pm$123.9 & 2.922 & 0.9763 & 96.68 & 1.35 & 0.00 & 0.072 \\
Static-MedRF & 1468.1$\pm$4.8 & 2.936 & 0.9906 & 99.65 & 1.52 & 10.24 & 1.013 \\
 
\bottomrule\end{tabular}
\end{table}

\begin{table}[!htbp]
\centering\scriptsize
\caption*{\textbf{Table S21.} R2 paired effects against Static-LightLR. Energy and adjusted-power intervals use ten within-block differences. F1 is percentage points; $p$ is the exact two-sided sign-flip diagnostic for energy.}
\setlength{\tabcolsep}{2.5pt}
\begin{tabular}{lrrrr}\toprule
Method & $\Delta E$ (J), 95\% CI & $\Delta P_{adj}$ (mW), 95\% CI & $\Delta$F1 & $p$\\\midrule
Cascade & +31.3 [-1.4,+63.9] & +98.8 [+77.9,+119.8] & +1.38 & 0.0762 \\
DeadlineGuard & +7.6 [+5.0,+10.1] & +14.2 [+0.4,+27.9] & +0.06 & 0.0020 \\
Tabular-Q & -15.9 [-49.9,+18.1] & -13.8 [-32.7,+5.1] & -23.29 & 0.2637 \\
DQN & +44.2 [-43.2,+131.7] & -30.8 [-128.1,+66.5] & +0.00 & 0.0020 \\
Static-MedRF & +51.3 [+49.5,+53.0] & +98.6 [+79.3,+117.9] & +1.43 & 0.0020 \\
 
\bottomrule\end{tabular}
\end{table}

\begin{table}[!htbp]
\centering\scriptsize
\caption*{\textbf{Table S22.} R2 state/action and service evidence. Occupancy totals 5,000 windows/method. Switch counts follow the implementation's logging convention: one initialization transition from the TinyDT placeholder $a_0$ is logged for every policy whose first action is not TinyDT, so static policies log 1.0 and runtime switches equal the logged value minus one for those policies.}
\setlength{\tabcolsep}{2.2pt}
\begin{tabular}{lrlrrrr}\toprule
Method & States & Action occupancy & Switches & Fallback & Misses & Backlog\\\midrule
Static-LightLR & 15.4 & LightLR:5000 & 1.0 & 0 & 0 & 0 \\
Cascade & 15.4 & ConfidenceCascade\_0p3\_0p7:5000 & 1.0 & 0 & 21 & 0 \\
DeadlineGuard & 15.4 & LightLR:1500, MedRF:3500 & 5.3 & 0 & 0 & 0 \\
Tabular-Q & 10.7 & TinyDT:5000 & 0.0 & 5000 & 0 & 0 \\
DQN & 15.6 & LightLR:5000 & 1.0 & 0 & 0 & 0 \\
Static-MedRF & 15.2 & MedRF:5000 & 1.0 & 0 & 512 & 0 \\
 
\bottomrule\end{tabular}
\end{table}

DeadlineGuard selects MedRF in 3,500 and LightLR in 1,500 windows and records 5.3 logged switches/run, i.e., 4.3 runtime switches/run after excluding the initialization transition. Against Static-MedRF it uses 43.714 J less per run [41.418,46.010], reduces deadline misses by 10.24 percentage points to zero, and loses 1.373 F1 pp. Frozen DQN remains LightLR. Frozen Tabular-Q selects TinyDT in all 5,000 windows. The registry calls these selections ``fallback'', but the frozen source contains no separate exception branch: unseen zero-initialized Q rows are tied, NumPy argmax chooses the first maximum, and the action order places TinyDT first. The R2 behavior is therefore the direct hardware realization of the off-support tie/default already visible in the 314/324 TinyDT full greedy map.

DQN's raw energy difference from Static-LightLR is positive in all ten blocks, with median +5.72 J, but Block 4 is +392.0 J and raises the SD of paired differences to 122.2 J; the Student-$t$ 95\% interval consequently spans $[-43.2,131.7]$ J. The same run has a 3.919-W preceding idle and 49.2$^\circ$C maximum temperature. Preceding-idle-adjusted power does not resolve a DQN--LightLR difference. We therefore retain both the 10/10 raw direction (exact sign-flip $p=0.00195$) and the unstable magnitude rather than collapsing them into an ``equivalent'' or causal controller-power claim.

\begin{figure}[!htbp]
\centering\includegraphics[width=.75\textwidth]{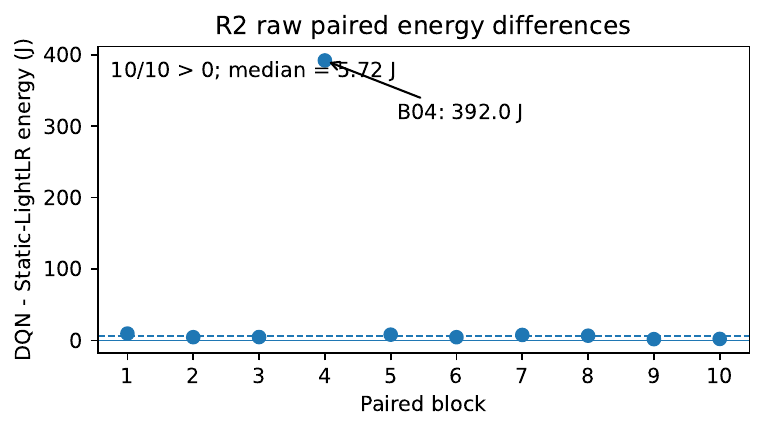}
\caption*{\textbf{Figure S8a.} R2 DQN-minus-Static-LightLR raw energy by paired block. Nine differences are between 1.8 and 9.6 J; Block 4 is a 392.0-J power/temperature excursion.}
\end{figure}

\begin{figure}[!htbp]
\centering\includegraphics[width=.95\textwidth]{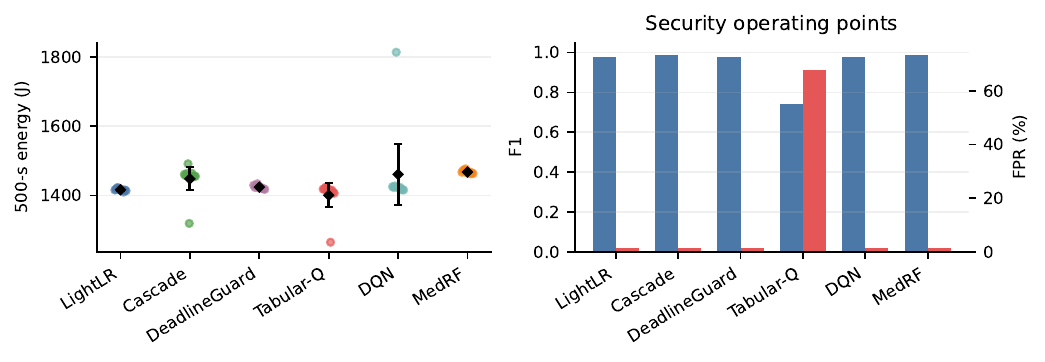}
\caption*{\textbf{Figure S8.} R2 raw energy distributions and security operating points. The block is the repeat unit.}
\end{figure}

\FloatBarrier
\section{R3 stop and R4 TinyDT threshold diagnostic}
R3 required four original observed-support states for controller-only whole-device measurement. The frozen source exposes eight and contains no outcome-independent rule for selecting four. The campaign therefore stopped before measurement. No controller-only power or joule value is inferred from P1B timing.

R4A selects threshold 0.6467505097 solely on validation balanced accuracy with the highest-threshold tie rule, freezes its hash, and applies it once to test. R4B is not executed because an independent stable reference load is unavailable.

\begin{table}[!htbp]
\centering\footnotesize
\caption*{\textbf{Table S23.} TinyDT test operating point at the primary and diagnostic thresholds. The diagnostic does not replace the primary.}
\setlength{\tabcolsep}{3.0pt}
\begin{tabular}{lrrrrrrrrr}\toprule
Result & $\tau$ & TN & FP & FN & TP & F1 & Rec. & FPR & BA\\\midrule
Primary fixed & 0.500 & 5519 & 11610 & 13 & 5348 & 0.4792 & 99.76 & 67.78 & 0.6599 \\ 
Validation-selected diagnostic & 0.646751 & 16817 & 312 & 21 & 5340 & 0.9698 & 99.61 & 1.82 & 0.9889 \\ 
 
\bottomrule\end{tabular}
\end{table}

The diagnostic changes test F1 from 0.4792 to 0.9698 and FPR from 67.78\% to 1.82\% while recall changes from 99.76\% to 99.61\%. Test AUROC is 0.9906. This supports calibration/threshold sensitivity as an explanation for much of the threshold-0.5 failure, but the analysis was introduced after the discrepancy was known and cannot repair frozen policy outcomes.

\begin{figure}[!htbp]
\centering\includegraphics[width=.95\textwidth]{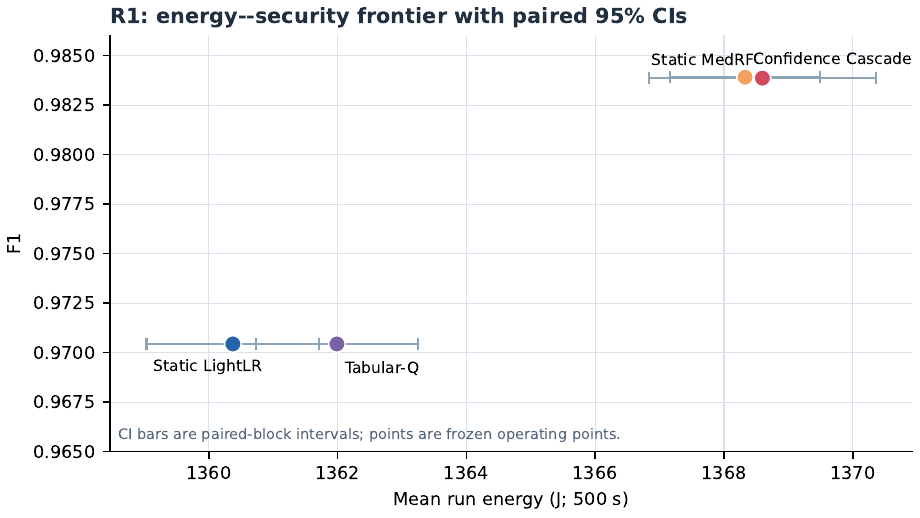}
\caption*{\textbf{Figure S9.} R1 energy--security frontier. Horizontal bars are paired-block 95\% CIs for run energy; the plotted F1 values are method means.}
\end{figure}

\begin{figure}[!htbp]
\centering\includegraphics[width=.95\textwidth]{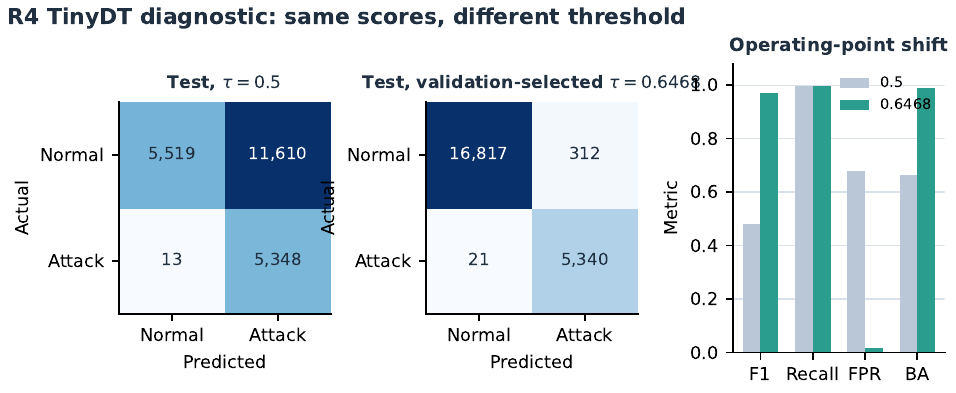}
\caption*{\textbf{Figure S10.} R4 TinyDT threshold diagnostic. The primary threshold remains 0.5; the validation-selected threshold is a post-hoc diagnostic only.}
\end{figure}

\FloatBarrier
\clearpage
\section{Final package integrity}
The P0/P1 evidence ZIP has SHA-256:
\begin{center}\small\texttt{d4e1e96960f1279c79445c744a02be1033e288dad27f440b87753934a3f4b108}\end{center}
Its size is 141,322,551 bytes and it contains 1,918 ZIP entries. Its root manifest contains 1,917 entries and independently covers every content file except itself; CRC, manifest, nested-manifest, and final validator checks pass. The follow-up (R0--R4) evidence ZIP has SHA-256:
\begin{center}\small\path{845ced5c96f932ee8d3509e2effcb39870a2fb9995bad9a6fe524f6c60b270e4}\end{center}
It has 1,912 entries and a 1,911-file root manifest. Its final validator reports R0/R1/R2/R4 PASS, R3 as a predeclared-ambiguity stop, no policy retraining, no test-driven selection, and retained invalid/replacement linkage. These properties establish internal provenance, not external validity.

\FloatBarrier
\endgroup

\end{document}